\RequirePackage{fix-cm}
\documentclass[letterpaper]{article}
\usepackage{microtype}
\usepackage{graphicx}
\usepackage[table]{xcolor}
\usepackage{booktabs}
\usepackage{tikz}
\usepackage{hyperref}

\usepackage[preprint]{icml2027}

\usepackage{amsmath,amssymb,amsthm}
\theoremstyle{definition}
\newtheorem{haartheorem}{Theorem}
\newtheorem{proposition}{Proposition}
\icmltitlerunning{Direction-Adaptive Haar Framelets on Graphs}
\hypersetup{hidelinks,pdfsubject={Preprint}}

\usepackage{etoolbox}
\makeatletter
\let\pedaf@correspondingauthor\icmlcorrespondingauthor
\renewcommand{\icmlcorrespondingauthor}[2]{%
  \ifstrempty{#2}
    {\gdef\icmlcorrespondingauthor@text{#1}}
    {\pedaf@correspondingauthor{#1}{#2}}}
\makeatother

\begin{document}
\twocolumn[
    \icmltitle{Direction-Adaptive Haar Framelets on Graphs}
\begin{icmlauthorlist}
    \icmlauthor{Zehua Fang}{cityuhk}
    \icmlauthor{Xiaosheng Zhuang}{cityuhk}
\end{icmlauthorlist}
\icmlaffiliation{cityuhk}{City University of Hong Kong}
\icmlcorrespondingauthor{Xiaosheng Zhuang}{}

    \icmlkeywords{Graph Neural Networks, Directional Wavelets, Haar Framelets}
    \vskip 0.3in
]
\printAffiliationsAndNotice{}
\raggedbottom

\begin{abstract}
Haar framelets on graphs extract local differences at different scales by comparing signals from children of the same parent, but building a tree from edge connectivity alone can assign adjacent nodes with markedly different features to different parents, making these differences hard to capture in the finest-scale high-pass responses.
Inspired by how classical directional wavelets analyze local variation in rotated coordinates, we propose direction-adaptive Haar framelets on graphs (DAHF), a framework that changes what Haar compares at each scale by adapting the partition tree.
DAHF learns candidate directions from feature differences between adjacent nodes and builds the partition tree level by level, favoring structurally admissible groupings in which feature variation within the parent is concentrated along one direction.
Following PEGFAN's Haar construction, the complete transform preserves the input signal's energy and supports exact reconstruction, and its node responses are permutation equivariant when the nodes and partition tree are relabeled together.
For two stripe signals on a periodic grid, the fraction of energy entering the finest-scale high-pass band rises from about \(43\%\)--\(47\%\) with PEGFAN using the Paris clustering method built into the authors' example code (Paris-PEGFAN) to \(85\%\)--\(86\%\) with DAHF.
Across nine node classification datasets, DAHF achieves an average accuracy of \(69.09\%\), ranking first among the methods compared and exceeding preset Paris-PEGFAN by \(0.75\) percentage points.
\end{abstract}

\section{Introduction}
\label{sec:introduction}

A graph consists of nodes and the edges connecting them. Graphs are commonly used to describe citation networks, social networks, and molecular structures. Nodes can have attributes, and edges record relationships between nodes. In node classification, a graph exhibits homophily when adjacent nodes tend to share a label, as when papers in the same research area cite one another. It exhibits heterophily when neighbors tend to belong to different classes. Graph convolutional networks (GCNs) learn node representations by aggregating neighbor features \citep{kipf2017semi}. On heterophilous graphs, one-hop neighbors can differ substantially from the central node \citep{luan2022revisiting}, and useful information can lie farther away. Some GNNs therefore process a node's own features and its neighbors' features separately and use multiple hops \citep{zhu2020beyond,maurya2021improving}. How can a model use features from neighborhoods of different sizes while preserving local contrasts?

One class of graph neural networks addresses this problem using graph wavelets and framelets \citep{zheng2021framelets,li2024permutation}. Classical wavelets and framelets in Euclidean spaces are typical examples of multiscale representations. They decompose signals into coarse approximations and details at different scales, preserving local variation while summarizing overall structure \citep{mallat1989multiresolution,unser2013unifying}. The same principle applies to a graph when its node features are treated as a signal. Low-pass components summarize features over larger regions, while high-pass components capture local differences. Graph wavelets can be constructed in the spectral domain of the graph Laplacian \citep{hammond2011wavelets,dong2017sparse} or directly in the vertex domain, as in PEGFAN \citep{li2024permutation}. In the original PEGFAN experiments, the graph edges determine a hierarchy of clusters, and the resulting partition tree supports a Haar framelet transform. At each parent, a normalized sum of its children's signals yields the low-pass component, and normalized pairwise differences yield the high-pass components. Tree levels determine the scales, and children sharing a parent form the neighborhood over which the transform compares signals. Appendix~\ref{app:related-work} reviews related work on graph wavelets, directional analysis, and hierarchical clustering.

Building a partition tree from edge connectivity alone may leave important local differences unrepresented in Haar high-pass responses at different scales. Our Paris-PEGFAN reference uses the Paris clustering method built into the PEGFAN authors' graph framelet example code \citep{bonald2018hierarchical}. We use this reference to analyze alternating column or row stripe signals on the same periodic grid. Pairing nodes along a stripe yields a zero Haar difference even when the neighboring stripe has the opposite value. For the two patterns, only \(43.4\%\) and \(46.7\%\) of the total energy of the Haar input enter the finest-scale high-pass band. The classifier may therefore receive fewer local contrasts from the finest-scale channel than are observable on the original graph. When these contrasts carry class information, it must extract them from other scales or propagation channels. This effect may also occur in other partition-tree Haar constructions. For example, the Haar-like wavelets of \citet{gavish2010multiscale} restrict each detail function's support to its corresponding parent. From the perspective of directional analysis, once the partition is fixed, the local neighborhoods in which wavelets compare signals are fixed as well. When the main signal variation crosses partition boundaries at a given scale, that scale's high-pass response may be weak.

Classical directional wavelets suggest a way to adapt these comparisons. At a fixed location, rotating a wavelet can align its analysis direction with the local signal variation and strengthen its response \citep{freeman1991steerable,unser2013unifying}. Neighborhoods on general graphs have no predefined spatial coordinates, so it is difficult to define such a rotation directly. In Euclidean space, however, the rotated wavelet \(\psi(R_\theta^{-1}u)\) can be viewed as the fixed mother wavelet \(\psi\) evaluated in rotated local coordinates. This observation leads us to propose DAHF, a method for constructing direction-adaptive partition trees. We realize \textquotedblleft{}rotation\textquotedblright{} on a graph by regrouping children. The Haar difference rule stays fixed, and the grouping determines which children it compares.
We first fit a set of directions to feature differences between adjacent nodes. We then screen candidate children by edge connectivity and favor groupings in which feature variation within the proposed parent is concentrated along one direction. Different parents can select different directions across locations and scales. We apply PEGFAN's Haar construction and classifier to the resulting tree, combining multiscale Haar responses with graph propagation features \citep{li2024permutation}. The complete Haar transform preserves signal energy and can reconstruct the input exactly from the coefficients across scales. Its node responses and the classifier outputs are permutation equivariant when the nodes and the given tree are relabeled together. With direction-adaptive partitions, the finest-scale high-pass energy shares for the two stripe signals rise to \(85.1\%\) and \(86.2\%\). On nine node classification benchmarks, DAHF attains a mean test accuracy of \(69.09\%\), the highest among the methods compared. Relative to preset Paris-PEGFAN, it improves mean accuracy by \(0.75\) percentage points and mean Macro-F1 by \(1.56\) percentage points.

We make three contributions.
\begin{enumerate}
    \item We interpret the partition tree as a set of multiscale neighborhoods for Haar analysis on graphs. Tree levels determine the analysis scale, and the composition of each parent determines which signals Haar compares at that scale. Based on this interpretation, we introduce a notion of local direction defined by the graph signal: the candidate axis that best explains feature variation among the children. We use this direction to guide grouping so that Haar neighborhoods can adapt to the direction of local variation.
    \item We propose an algorithm for building a direction-adaptive partition tree. At each scale, it uses graph connectivity to screen candidates and selects groupings based on feature variation across the entire proposed parent. With PEGFAN's Haar construction, the complete transform supports exact reconstruction and preserves signal energy. It is permutation equivariant when the nodes and partition tree are relabeled together.
    \item We integrate the direction-adaptive partition tree into PEGFAN and compare it with fourteen tuned baselines and preset Paris-PEGFAN on nine node classification benchmarks using the same splits and metrics. DAHF achieves the highest nine-dataset average accuracy in this comparison, at \(69.09\%\). Relative to the Paris-based reproduction of PEGFAN, average accuracy and average Macro-F1 increase by \(0.75\) and \(1.56\) percentage points, respectively.
\end{enumerate}

\section{Constructing Rotated Wavelets on Graphs}
\label{sec:method}

\subsection{From Two-Dimensional Directional Wavelets to Haar Analysis on Graphs}

A wavelet measures local signal variation at a specified position and scale \citep{mallat1989multiresolution}. Haar analysis uses a normalized sum for the low-pass component and normalized differences for the high-pass components. Within an image patch, the former summarizes intensity and the latter compares two subregions. Appendix~\ref{app:wavelet-background} introduces the Haar wavelet and explains how these operations analyze a signal level by level.

\begin{figure}[t]
    \centering
    \includegraphics[width=\columnwidth]{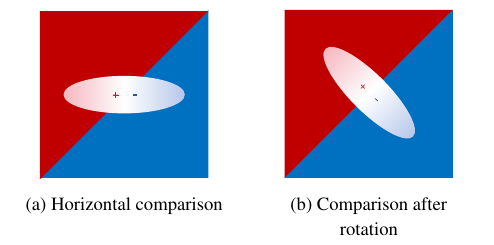}
    \caption{Directional wavelets at a \(45^\circ\) intensity boundary. The ellipse represents the wavelet support, and the red-to-blue gradient inside it indicates the positive and negative weight regions. Before rotation, both weight regions cross the intensity boundary. After rotation, they lie on opposite sides, and the difference directly reflects the intensity change.}
    \label{fig:diagonal-haar}
\end{figure}

Suppose an image contains a boundary at \(45^\circ\) to the horizontal, with a bright region on one side and a dark region on the other. In a left--right comparison, both subregions may contain bright and dark pixels, weakening their contrast after averaging. Rotating the wavelet so that its positive and negative weights fall on opposite sides of the boundary makes the difference directly reflect the intensity jump. Directional wavelets can therefore reveal local variation more clearly at the same position and scale by changing direction \citep{unser2013unifying,xiao2021adaptive}. Figure~\ref{fig:diagonal-haar} compares these two cases.

For a single wavelet, rotating its support and the associated weight distribution is equivalent to keeping the wavelet fixed and rotating the local coordinate axes in the opposite direction.

\begin{proposition}[Wavelet rotation and inverse rotation of local coordinates]
\label{prop:wavelet-rotation}
Let \(f,\psi\in L^2(\mathbb R^2)\), where \(\psi\) is a wavelet at a fixed scale. With analysis center \(b\), define \(f_b(u)=f(b+u)\). Let \(R_\theta\) be a rotation matrix and \(\psi_\theta(u)=\psi(R_\theta^{-1}u)\) the rotated wavelet. Then
\[
    \langle f_b,\psi_\theta\rangle
    =\langle f_b\circ R_\theta,\psi\rangle.
\]
\end{proposition}

Appendix~\ref{app:wavelet-rotation-proof} proves Proposition~\ref{prop:wavelet-rotation}.

This equivalence allows us to keep the wavelet fixed and use the local coordinates \(v=R_\theta^{-1}u\) to obtain the same coefficient as wavelet rotation and adjust the regions being compared.

On graphs, PEGFAN uses a partition tree to organize analysis neighborhoods at different scales \citep{li2024permutation}. The original nodes are the leaves at the finest level. They first form small clusters, which are then grouped into larger parents level by level. As clusters contain more nodes, the analysis neighborhoods grow. As in two-dimensional Haar analysis, each parent computes a low-pass signal summarizing the overall level and high-pass signals recording local differences. For child signals \(y_1,\ldots,y_p\), the low-pass signal is the normalized sum \(p^{-1/2}\sum_{i=1}^{p}y_i\), and the high-pass signals are normalized differences \(p^{-1/2}(y_i-y_j)\) for \(i<j\). The low-pass signal becomes the input to the next level, while the high-pass signals remain at the current level.

The original PEGFAN experiments build the tree using graph edges alone \citep{li2024permutation}. Grouping solely by connectivity may prevent important local differences from appearing in the high-pass responses at the corresponding scales. We place alternating stripe signals on a periodic grid and examine this effect using a Paris-based reproduction of PEGFAN. Pairing nodes along the same stripe yields a zero finest-scale Haar difference even when the neighboring stripe has the opposite value. For alternating column and row stripes, the finest-scale high-pass band captures only \(43.4\%\) and \(46.7\%\) of the energy of the Haar input, respectively (Figure~\ref{fig:grid-experiment}).

Inspired by coordinate rotation in Euclidean space, we adjust local graph Haar analysis neighborhoods through regrouping. Specifically, among groupings allowed by the graph structure, we favor placing children whose feature differences lie mainly along the same direction into one parent. Haar differences between these children can then reflect feature variation along that direction.

\begin{figure}[t]
    \centering
    \includegraphics[width=.98\columnwidth]{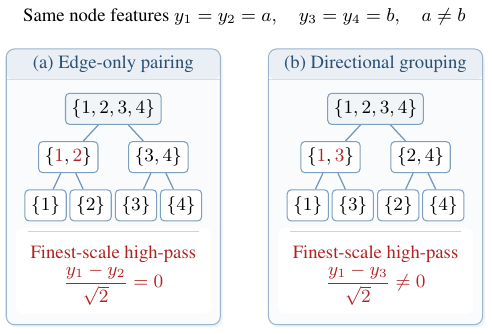}
    \caption{Regrouping aligns the finest-scale Haar comparison with feature variation. Red digits mark the cluster containing node \(1\), and the high-pass response is also highlighted in red. Left: pairing by edges alone groups nodes \(1\) and \(2\), whose identical features yield a zero difference. Right: regrouping by feature direction yields a high-pass response along \(a-b\) for nodes \(1\) and \(3\).}
    \label{fig:haar-regrouping}
\end{figure}

Figure~\ref{fig:haar-regrouping} illustrates this alignment with four nodes. In this example, pairing by edges alone places nodes \(1\) and \(2\) in the same parent. Their identical features yield a zero finest-scale high-pass response. Grouping nodes \(1\) and \(3\) instead lets the same Haar difference rule detect variation along \(y_1-y_3\). As in Figure~\ref{fig:diagonal-haar}, adjusting which signals are compared allows the difference to reflect variation across the boundary more clearly.

\subsection{Constructing a Direction-Adaptive Partition Tree}

We first fit candidate directions from the original node features as candidate rotation directions for graph wavelets. We then analyze Paris clustering and modify its grouping procedure to accommodate directional scoring. We construct a layerwise grouping procedure using only edge information, then add directional scoring within the same procedure.

\subsubsection{Fitting Candidate Directions}
\label{sec:candidate-directions}

For each node, we first sum all feature entries. If the sum is nonzero, we divide every entry by it. If the sum is zero, we retain the original feature vector. Let \(x_u\) denote the normalized feature vector of node \(u\). We sample undirected edges from the original graph after removing self-loops and compute \(x_u-x_v\) for each sampled edge \((u,v)\). We discard near-zero differences and normalize the remaining differences to unit length.

Each normalized difference is a sample, assigned to the current axis with the largest absolute inner product. For samples assigned to the same axis, we align their signs, sum them, and normalize the sum to update the axis. The absolute inner product treats \(q\) and \(-q\) as the same axis, so exchanging the endpoints of an edge does not change its direction-matching scores.
The resulting dictionary is \(Q=\{q_1,\ldots,q_K\}\), where \(K\leq R\) and \(R\) is the preset direction limit. We fit the dictionary once from node features and keep it fixed during tree construction and classifier training. Appendix~\ref{app:direction-dictionary} gives the edge sampling, axis initialization, and update details.

\paragraph{Why change the Paris grouping procedure?}
We aim to incorporate directions of feature variation while retaining the original graph's edge-connectivity constraints. This corresponds to rotating a wavelet support within a given analysis neighborhood in Euclidean space to align it with local variation. However, the merge order in Paris is determined by intercluster distances derived from the adjacency matrix, and its nondecreasing merge heights depend on the original distance update formula \citep{bonald2018hierarchical}. Directly adding a direction weight to pairwise similarity can make a later merge height lower than an earlier one, violating this ordering. Appendix~\ref{app:paris-angle-counterexample} gives a three-node counterexample.

We therefore use the Paris structural score while modifying the underlying grouping procedure to construct a partition tree that follows edge-connectivity constraints and accommodates directional scoring.

\subsubsection{Edge-Based Layerwise Grouping}
\label{sec:parent-construction}

The level-\(0\) vertices are the original nodes. At higher levels, each vertex is a cluster formed at the previous level. Let \(V_i\) be the set of original nodes in current-level cluster \(i\). If \(A\) is the original adjacency matrix with self-loops removed, the edge weight between distinct level-\(\ell\) clusters \(i,j\) is
\[
    A_{ij}^{(\ell)}=\sum_{u\in V_i}\sum_{v\in V_j}A_{uv},
    \qquad i\ne j.
\]
The diagonal of \(A^{(\ell)}\) is zero. At level \(0\), each \(V_i\) contains one original node, so \(A^{(0)}=A\).

\paragraph{Filtering candidates by graph structure.}
We group level-\(\ell\) children into a parent that becomes a cluster at level \(\ell+1\). Let \(S\) be the set of level-\(\ell\) children already selected for this parent. An unassigned cluster \(j\) at the same level is a candidate for joining \(S\) if it has a positive-weight edge to at least one child in \(S\). Denote this candidate set by \(\mathcal J(S)\).

On the level-\(\ell\) graph, cluster \(i\) has weighted degree \(v_i=\sum_k A_{ik}^{(\ell)}\), where \(k\) ranges over all clusters at that level. We compute \(v_S=\sum_{i\in S}v_i\) and \(T_\ell=\sum_i v_i\). The former sums only over the children in \(S\), and the latter over all level-\(\ell\) clusters. The candidate degree \(v_j\) is computed in the same way at the same level. We use the reciprocal of a Paris-style distance as the structural score \citep{bonald2018hierarchical}:
\begin{equation}
    P_\ell(S,j)=\frac{T_\ell\,w(S,j)}{v_Sv_j},
    \qquad
    w(S,j)=\sum_{i\in S}A_{ij}^{(\ell)}.
    \label{eq:structural-score}
\end{equation}

\paragraph{Choosing a seed pair and growing the parent.}
For a seed pair, candidates are positive-weight edges \((i,j)\) whose endpoints are both unassigned. We score each edge by \(P_\ell(\{i\},j)\), choose the edge with the highest structural score as the seed \(S=\{i,j\}\), and mark both children assigned. We then select the highest-scoring \(j\) in \(\mathcal J(S)\), add it to \(S\), and mark it assigned. After each addition, we update and rescore the candidates. We stop when \(S\) has \(h\geq2\) children or no positive-weight edge connects \(S\) to an unassigned cluster. For ties in structural score, seed selection prefers the larger edge weight and then orders by endpoint indices. Growth ties are resolved by candidate index.

Let \(P_{\max}\) be the highest structural score at each step. We select only candidates satisfying \(P=P_{\max}\). We call this version \textbf{edge-based layerwise grouping}.

\subsubsection{Direction-Adaptive Grouping}
\label{sec:directional-grouping}

To account for both feature-variation directions and graph edges, we relax the selection condition \(P=P_{\max}\) to \(P\geq\tau P_{\max}\), where \(\tau\in[0,1]\). All candidates satisfying this condition are retained, and directional scoring selects among them.

\paragraph{Scoring a proposed parent's direction.}
For each retained candidate \(j\), let \(U_j=S\cup\{j\}\) be the set of children formed by adding \(j\). We evaluate this grouping by the fraction of feature variation within \(U_j\) concentrated along one candidate axis. Since Haar compares children sharing a parent, the score considers both the new candidate and the selected children.

For each child \(i\), let \(m_i=|V_i|\) be its number of original nodes and \(\mu_i=m_i^{-1}\sum_{u\in V_i}x_u\) their mean feature vector. At level \(0\), \(m_i=1\) and \(\mu_i=x_i\). For the child set \(U\) of any proposed parent, the weighted mean is \(\bar\mu_U=(\sum_{i\in U}m_i\mu_i)/(\sum_{i\in U}m_i)\). Total variation about this mean and variation along \(q_r\) are
\begin{align}
    E(U)&=\sum_{i\in U}m_i\|\mu_i-\bar\mu_U\|_2^2,\\
    E_r(U)&=\sum_{i\in U}m_i\bigl[q_r^\top(\mu_i-\bar\mu_U)\bigr]^2.
\end{align}
We find the dictionary axis that explains the most variation and use its explained fraction as the directional score:
\begin{equation}
    C(U)=
    \left\{
    \begin{alignedat}{2}
        &\displaystyle\max_{1\leq r\leq K}\frac{E_r(U)}{E(U)},\;&&K>0,\ E(U)>10^{-20},\\[3pt]
        &0,&&\text{otherwise}.
    \end{alignedat}
    \right.
    \label{eq:directional-score}
\end{equation}
Different parents can select different axes, so directions within one tree can vary with position and scale.

\begin{figure*}[t]
    \centering
    \includegraphics[width=.95\linewidth]{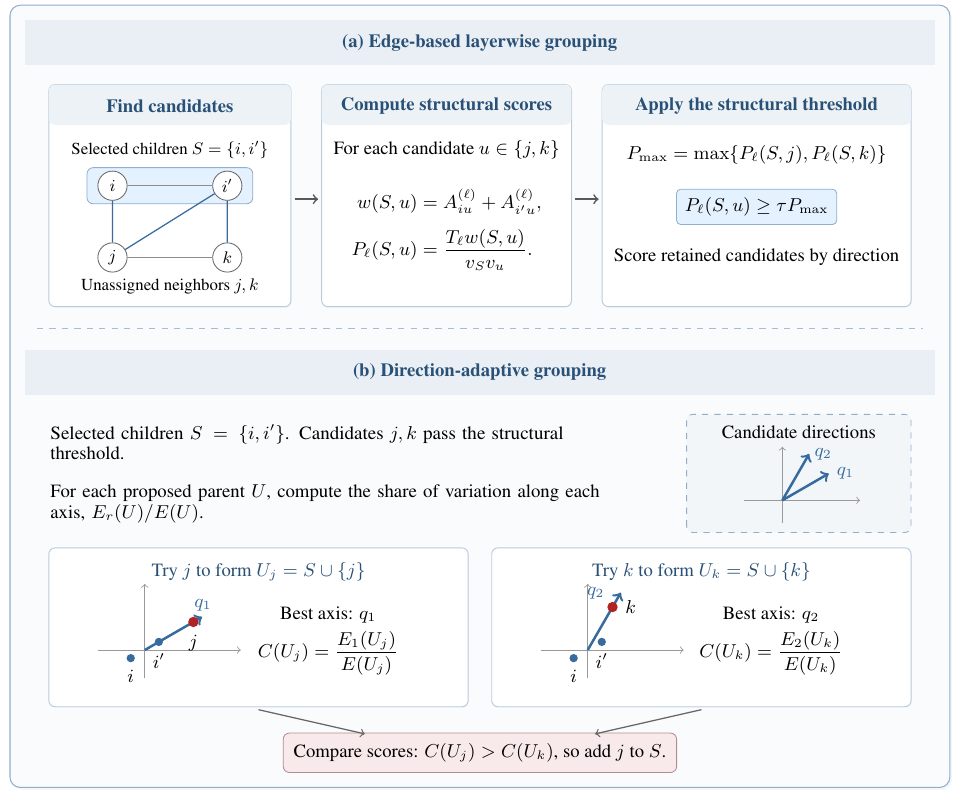}
    \caption{One step of growing a level-\((\ell+1)\) parent. (a) Edge-based layerwise grouping. The selected children inside the box and the candidates connected by blue edges are all at level \(\ell\). (b) Direction-adaptive grouping. The upper inset shows two dictionary directions. The lower panels use the same feature coordinates. Blue dots indicate selected children, red dots indicate candidates being considered, and blue arrows show the highest-scoring direction for each grouping. Comparing the two scores leads to adding \(j\) to \(S\).}
    \label{fig:parent-construction}
\end{figure*}

\paragraph{Selecting among candidates.}
Among structurally retained candidates, selection follows two rules. \textbf{Seed selection:} maximize \(C(\{i,j\})\); break ties by the larger structural score, then the larger edge weight, then the smaller endpoint indices (first \(i\), then \(j\), with \(i<j\)). \textbf{Parent growth:} maximize \(C(S\cup\{j\})\); break ties by the larger structural score, then the smaller candidate index. Each later criterion is used only when all earlier criteria tie. Figure~\ref{fig:parent-construction} shows one expansion step. We call this version \textbf{direction-adaptive grouping}.

Greedily selecting the current highest-scoring candidate can miss a child combination with a higher final directional score. To measure this gap, we compare the first parent produced by the greedy method with the highest-scoring parent reachable by exhaustive search under the same structural filtering rule in \(1{,}080\) local neighborhoods across nine datasets. Exhaustive search improves \(C\) by only \(0.0084\) on average. Appendix~\ref{app:greedy-audit} gives the setup and per-dataset results.

\subsubsection{Complete Tree Construction and Hierarchical Ordering}

Starting from the original nodes, we repeatedly form parents at every level. Both the edge-based and direction-adaptive versions use this procedure, differing only in candidate selection as described in Sections~\ref{sec:parent-construction} and \ref{sec:directional-grouping}. Each parent becomes a vertex at the next level. Edge weights in the new graph sum the weights between distinct parents, and parent features average child features weighted by their numbers of original nodes. Each filtering step recomputes \(P_{\max}\) from its current candidates. If grouping stops because no edges remain, the final root can have more than \(h\) children. Appendix~\ref{app:tree-algorithm} gives the complete pseudocode in Algorithm~\ref{alg:direction-tree}.

The following theorem guarantees that the algorithm constructs a partition tree with a strict hierarchical ordering while accounting for both directional information and edge-connectivity constraints.

\begin{samepage}
\begin{haartheorem}[Candidate selection and hierarchical ordering of the partition tree]
\label{thm:candidate-order}
Fix the current partition state, direction dictionary, and cluster indices. If the candidate set for seed selection or parent growth is nonempty, the structurally retained candidates form a strict total order under the rules above, so the preferred candidate is unique. Its structural score \(P_{\mathrm{sel}}\) satisfies \(P_{\mathrm{sel}}\geq\tau P_{\max}\). In particular, if \(\tau=1\) and the highest structural score is unique, this step is determined solely by the structural score.

Algorithm~\ref{alg:direction-tree} produces a partition tree with a unique root in finitely many steps. Each coarsening step merges the current partition into a coarser partition. With level indices as cluster heights, every child and its parent satisfy
\[
    \operatorname{level}(\text{parent})
    =\operatorname{level}(\text{child})+1.
\]
\end{haartheorem}
\end{samepage}

Appendix~\ref{app:ordering} proves Theorem~\ref{thm:candidate-order}.

Thus, heights strictly increase along every path from a leaf to the root. Here, tree heights are defined by level indices, while the scores used to compare candidates may change as a parent grows.

\subsection{Haar Transform on the Partition Tree and Its Properties}
\label{sec:haar-transform}

Let \(Z\) be the input node signal. Following PEGFAN's Haar construction \citep{li2024permutation}, each parent computes a normalized sum and normalized pairwise differences of its children's signals. The low-pass signal passes upward, and the high-pass signals remain at the current scale. Let \(\Phi\) be the root low-pass operator, \(W_\ell\) the high-pass operator for the level-\((\ell+1)\) parents, and \(L\) the root level. After mapping coefficients back to the original nodes, the classifier uses the channels
\begin{equation}
    H_\ell=W_\ell^\top W_\ell Z,\qquad
    H_{\mathrm{low}}=\Phi^\top\Phi Z.
    \label{eq:haar-responses}
\end{equation}
Appendix~\ref{app:haar-operators} gives the operator recursion and proofs of the following properties.

\begin{haartheorem}[Energy preservation]
\label{thm:haar-energy}
The energy of the root low-pass coefficients plus the high-pass coefficient energies at all scales equals the input energy:
\[
    \|\Phi Z\|_F^2+\sum_{\ell=0}^{L-1}\|W_\ell Z\|_F^2=\|Z\|_F^2.
\]
\end{haartheorem}

\begin{haartheorem}[Exact reconstruction]
\label{thm:haar-reconstruction}
Summing the root low-pass response and all high-pass responses exactly reconstructs \(Z=H_{\mathrm{low}}+\sum_{\ell=0}^{L-1}H_\ell\).
\end{haartheorem}

\begin{haartheorem}[Sparsity]
\label{thm:haar-sparsity}
Each high-pass wavelet involves only a pair of children sharing one parent. For a single-channel input \(z\), let \(\widehat z\) collect all retained coefficients and \(h_\ell\) be the maximum number of children of a level-\((\ell+1)\) parent. Then
\[
    \|\widehat z\|_0\leq
    \left(1+\sum_{\ell=0}^{L-1}(h_\ell-1)\right)\|z\|_0,
\]
where \(\|\cdot\|_0\) counts nonzero entries.
\end{haartheorem}

\paragraph{Permutation equivariance.}
When the original nodes and their cluster memberships in the partition tree are relabeled together, the Haar node responses at every scale follow the same node permutation.

\subsection{Node Classification}

We use the PEGFAN Type~c setting \citep{li2024permutation}. Let \(X\) be the normalized original node features and \(\widehat A\) the symmetrically normalized adjacency. Haar node responses are computed from the once-propagated features \(Z=\widehat A X\). The classifier receives two types of channels. One consists of \(X\) and the three propagated features \(\widehat A X,\widehat A^2X,\widehat A^3X\). The other consists of the root low-pass response \(H_{\mathrm{low}}\) and the high-pass responses \(H_\ell\) at each level. The dictionary and tree are still constructed from \(X\) before propagation.

The classifier follows PEGFAN's design based on FSGNN \citep{li2024permutation,maurya2021improving}. Each channel first receives a separate linear transformation and bias, followed by optional \(\ell_2\) normalization of each node's output vector. Learned softmax weights then weight the channels, which are concatenated along the feature dimension and passed through ReLU, dropout, and a final linear layer for class scores. At inference, the same computation applies to every node, so permuting all input channels by the same node permutation permutes the classification outputs accordingly.

Appendix~\ref{app:computational-cost} gives the time and space complexity and measured runtime.

\section{Experiments}

\begin{table*}[!t]
    \centering
    \scalebox{.94}{%
        \begin{minipage}{\linewidth}
        \centering
        \renewcommand{\tablename}{Table}
        \setlength{\belowcaptionskip}{4pt}
        \captionsetup{font=normalsize,labelfont={it,normalsize}}
        \caption{\textbf{Node classification on nine datasets.} We report test accuracy (\%).}
        \label{tab:accuracy}
        \normalsize
        \setlength{\tabcolsep}{1.0pt}
        \renewcommand{\arraystretch}{1.0}
        \definecolor{AccuracyHighlight}{HTML}{E5B5B5}
        \definecolor{AccuracySecond}{HTML}{C5E9E3}
        \definecolor{AccuracyThird}{HTML}{AFC3D8}
        \newlength{\accuracyModelWidth}
        \settowidth{\accuracyModelWidth}{Paris-PEGFAN}
        \addtolength{\accuracyModelWidth}{1.8pt}
        \newcolumntype{Q}{>{\centering\arraybackslash}m{\dimexpr(\linewidth-\accuracyModelWidth-24\tabcolsep-4\arrayrulewidth)/11\relax}}
        \newcommand{\tablerankmark}[2]{\begingroup\setlength{\fboxsep}{0.9pt}\colorbox{#1}{#2}\endgroup}
        \newcommand{\tablemetric}[1]{\textbf{#1}}
        \newcommand{\tableheadingstrut}{\rule[-3.3pt]{0pt}{11.5pt}}
        \newcommand{\tablemodelgroup}[1]{%
            \multicolumn{12}{c}{\tableheadingstrut\bfseries\itshape #1}}
        \arrayrulecolor{black}
        \rowcolors{3}{black!2}{white}
        \begin{tabular}{l|QQQ|QQQ|QQQ|QQ}
            \toprule
            \tableheadingstrut & \multicolumn{3}{c|}{\textbf{Citation Networks}}
            & \multicolumn{3}{c|}{\textbf{WebKB}}
            & \multicolumn{3}{c|}{\textbf{Other Graphs}}
            & \multicolumn{2}{c}{\textbf{Overall}} \\
            \cmidrule(lr){2-4}\cmidrule(lr){5-7}\cmidrule(lr){8-10}\cmidrule(lr){11-12}
            \rowcolor{white}
            \tableheadingstrut\textbf{Model} & \tablemetric{Cora} & \tablemetric{CiteSeer} & \tablemetric{PubMed}
            & \tablemetric{Texas} & \tablemetric{Wis.} & \tablemetric{Cornell}
            & \tablemetric{Actor} & \tablemetric{Cham.-f} & \tablemetric{Squir.-f}
            & \tablemetric{Avg.} & \tablemetric{Rank} \\
            \midrule
            \rowcolor{black!6}
            \tablemodelgroup{MLP and GCN} \\
            GCN & 87.04 & 76.50 & 88.14 & 63.78 & 61.37 & 50.54 & 30.69 & \tablerankmark{AccuracyHighlight}{42.90} & \tablerankmark{AccuracyHighlight}{42.32} & 60.36 & 16 \\
            MLP & 75.55 & 73.41 & 87.81 & 81.35 & 84.12 & 74.86 & 35.32 & 28.71 & 35.05 & 64.02 & 12 \\
            \midrule[0.35pt]
            \rowcolor{black!6}
            \tablemodelgroup{Feature Propagation and Fusion} \\
            GAMMA & 86.44 & 75.95 & 89.23 & 64.32 & 78.24 & 50.00 & 35.43 & 37.17 & 34.87 & 61.29 & 15 \\
            ACM-GCN++ & 84.59 & 73.84 & 87.11 & 73.24 & 71.18 & 63.78 & 34.09 & 39.71 & 38.39 & 62.88 & 13 \\
            ACM-GCN & 87.32 & 76.89 & \tablerankmark{AccuracyThird}{89.68} & \tablerankmark{AccuracySecond}{85.68} & 85.10 & \tablerankmark{AccuracyThird}{78.11} & 35.32 & 36.69 & 32.48 & 67.47 & 6 \\
            FSGNN & 87.42 & 76.87 & 89.61 & 83.78 & 86.08 & 76.49 & \tablerankmark{AccuracySecond}{36.57} & \tablerankmark{AccuracySecond}{42.65} & \tablerankmark{AccuracySecond}{41.61} & \tablerankmark{AccuracySecond}{69.01} & 2\textsuperscript{nd} \\
            \midrule[0.35pt]
            \rowcolor{black!6}
            \tablemodelgroup{Graph Filtering and Wave Propagation} \\
            AdaSpec & \tablerankmark{AccuracyThird}{87.81} & 74.78 & 87.67 & 68.38 & 64.51 & 62.16 & 34.41 & 36.09 & 35.90 & 61.30 & 14 \\
            OptBasisGNN & 85.31 & 74.24 & \tablerankmark{AccuracyHighlight}{89.83} & 74.86 & 75.88 & 65.14 & 35.36 & \tablerankmark{AccuracyThird}{41.88} & 35.68 & 64.24 & 11 \\
            ChebNetII & 87.44 & 76.58 & 89.43 & 80.81 & 81.57 & 77.57 & 35.78 & 38.15 & 33.96 & 66.81 & 9 \\
            GWN-fa & 87.40 & 76.69 & 89.28 & 82.16 & 84.71 & 75.14 & 35.76 & 32.41 & 39.54 & 67.01 & 8 \\
            CPF & \tablerankmark{AccuracySecond}{88.03} & \tablerankmark{AccuracyThird}{77.10} & 89.28 & \tablerankmark{AccuracyHighlight}{85.95} & 82.94 & \tablerankmark{AccuracySecond}{78.65} & 35.08 & 32.24 & 34.32 & 67.07 & 7 \\
            CASTOR & 87.22 & \tablerankmark{AccuracySecond}{77.36} & 89.22 & 82.16 & \tablerankmark{AccuracyHighlight}{87.65} & \tablerankmark{AccuracyHighlight}{80.54} & 35.26 & 38.06 & 36.14 & 68.18 & 4 \\
            \midrule[0.35pt]
            \rowcolor{black!6}
            \tablemodelgroup{Global Information Aggregation} \\
            SGFormer & 81.87 & 74.59 & 88.40 & 82.97 & 84.51 & 73.51 & \tablerankmark{AccuracyThird}{35.79} & 32.23 & 36.67 & 65.62 & 10 \\
            GloGNN++ & \tablerankmark{AccuracyHighlight}{88.15} & 77.01 & 89.31 & 85.14 & 86.27 & 74.05 & \tablerankmark{AccuracyHighlight}{37.45} & 37.95 & 34.92 & 67.81 & 5 \\
            \midrule[0.35pt]
            \rowcolor{black!6}
            \tablemodelgroup{Haar Framelets} \\
            Paris-PEGFAN & 86.30 & 76.72 & 89.37 & 85.14 & \tablerankmark{AccuracyThird}{87.45} & 76.76 & 34.54 & 40.31 & 38.48 & \tablerankmark{AccuracyThird}{68.34} & 3\textsuperscript{rd} \\
            \tablerankmark{AccuracySecond}{\textbf{(ours) DAHF}} & 87.22 & \tablerankmark{AccuracyHighlight}{77.51} & \tablerankmark{AccuracySecond}{89.75} & \tablerankmark{AccuracyThird}{85.41} & \tablerankmark{AccuracyHighlight}{87.65} & 77.84 & 35.23 & 41.31 & \tablerankmark{AccuracyThird}{39.92} & \tablerankmark{AccuracyHighlight}{69.09} & \textbf{1\textsuperscript{st}} \\
            \bottomrule
        \end{tabular}
        \arrayrulecolor{black}
        \par\smallskip
        \begin{minipage}{\linewidth}
            \normalsize\raggedright
            Entries average ten fixed splits (seed 42). Configurations maximize mean validation accuracy; checkpoints minimize validation loss. Tuned models try 16 candidates (15 complete candidates for AdaSpec/CiteSeer); Paris-PEGFAN uses its preset. DAHF fixes \(R=7\). Training and selection details are in Appendix~\ref{app:pedaf-selection}.

            Cham.-f/Squir.-f are filtered datasets. Pink/teal/light blue mark the top three per column, with ties sharing colors. Avg. weights datasets equally; Rank uses unrounded Avg. The 14 tuned baseline averages use the supplied two-decimal dataset means.
        \end{minipage}
        \end{minipage}%
    }
\end{table*}

\subsection{Experimental Setup}

\paragraph{Data and protocol.}
We evaluate nine graphs using the ten fixed Geom-GCN splits \citep{pei2020geom} and the filtered Chameleon/Squirrel splits \citep{platonov2023critical}, in a transductive setting with shared graph inputs. DAHF and fourteen tuned baselines each search sixteen configurations per dataset; Paris-PEGFAN uses its preset. We report test accuracy and Macro-F1 averaged over splits and weight datasets equally. Appendix~\ref{app:pedaf-selection} gives the full protocol and selection exception.

\begin{figure}[t]
    \centering
    \includegraphics[width=\columnwidth]{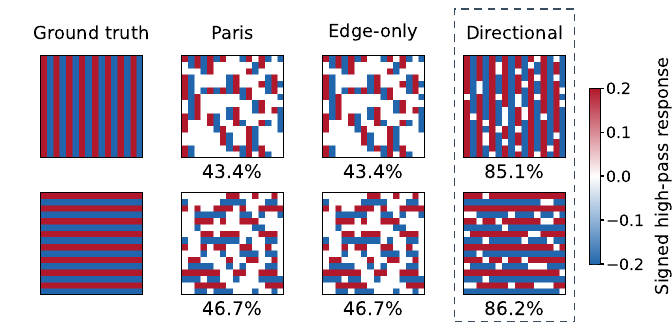}
    \caption{Finest-scale Haar responses. Ground truth is the input signal; the other columns share the signed-response color scale. Heatmaps share one relabeling, and percentages average energy shares over ten relabelings. See Appendix~\ref{app:grid-experiment}.}
    \label{fig:grid-experiment}
\end{figure}

\paragraph{Synthetic stripes.}
DAHF places \(85\%\)--\(86\%\) of the input energy in the finest-scale high-pass band, versus \(43\%\)--\(47\%\) for Paris-PEGFAN and edge-based grouping (Figure~\ref{fig:grid-experiment}; Appendix~\ref{app:grid-experiment}). More cross-stripe contrasts enter the finest-scale channel, while the complete Haar decomposition still preserves total energy.

\subsection{Node Classification on Real-World Graphs}

Table~\ref{tab:accuracy} ranks DAHF first among sixteen methods in mean accuracy (\(69.09\%\)), \(0.08\) percentage points above FSGNN, and second in Macro-F1 (\(58.69\%\)), \(0.96\) points below FSGNN. Against Paris-PEGFAN, its mean gains are \(0.75\) accuracy points and \(1.56\) Macro-F1 points (Appendix Table~\ref{tab:pedaf-paris}). GCN leads on both metrics on Chameleon-filtered and Squirrel-filtered. These rankings describe observed means.

\subsection{Hyperparameter Sensitivity}
\label{sec:tau-sensitivity}

The configurations selected over \(\tau\) and \(h\) vary by dataset. Cora and Squirrel-filtered both select \(h=8\) and \(\tau=0.30\). At fixed \(h=8\), these configurations exceed \(\tau=1\) by \(0.23\) and \(1.08\) validation accuracy points, respectively. Appendix~\ref{app:tau-sensitivity} reports all nine sensitivity curves.

\subsection{Ablation Study}
\label{sec:ablations}

\begin{table*}[!t]
    \centering
    \setlength{\belowcaptionskip}{6pt}
    \caption{\textbf{Partition-tree ablation.} Mean test accuracy (\%) over ten splits. Red values are paired gains over Paris-PEGFAN (pp). Avg. weights datasets equally. Setup: Appendix~\ref{app:structure-only-ablation}.}
    \label{tab:structure-accuracy}
    \footnotesize
    \setlength{\tabcolsep}{2.3pt}
    \renewcommand{\arraystretch}{1.24}
    \definecolor{AblationHighlight}{HTML}{D7E1E8}
    \newcommand{\ablationheader}[1]{\makebox[45pt][c]{\scriptsize #1}}
    \newcommand{\ablationcell}[2]{%
        \makebox[25pt][r]{#1}\hspace{1.5pt}%
        \makebox[18.5pt][l]{\fontsize{6}{7}\selectfont\bfseries\boldmath\textcolor[HTML]{D00000}{#2}}}
    \arrayrulecolor{black!38}
    \resizebox{\linewidth}{!}{\begin{tabular}{l|*{9}{c}|c}
        \toprule
        \rowcolor{black!12}
        \textbf{Model} & \ablationheader{Cora} & \ablationheader{CiteSeer} & \ablationheader{PubMed}
        & \ablationheader{Texas} & \ablationheader{Wis.} & \ablationheader{Cornell}
        & \ablationheader{Actor} & \ablationheader{Cham.-f} & \ablationheader{Squir.-f}
        & \ablationheader{\textbf{Avg.}} \\
        \midrule
        Paris-PEGFAN & \ablationcell{86.30}{} & \ablationcell{76.72}{} & \ablationcell{89.37}{} & \ablationcell{85.14}{} & \ablationcell{87.45}{} & \ablationcell{76.76}{} & \ablationcell{34.54}{} & \ablationcell{40.31}{} & \ablationcell{38.48}{}
        & \ablationcell{68.34}{} \\
        \rowcolor{black!4}
        Edge-based layerwise grouping
        & \ablationcell{86.92}{+0.62} & \ablationcell{76.54}{\ensuremath{-}0.18} & \ablationcell{89.80}{+0.43} & \ablationcell{85.68}{+0.54} & \ablationcell{87.65}{+0.20} & \ablationcell{77.30}{+0.54} & \ablationcell{34.31}{\ensuremath{-}0.23} & \ablationcell{39.80}{\ensuremath{-}0.52} & \ablationcell{38.99}{+0.51}
        & \ablationcell{68.55}{+0.21} \\
        \rowcolor{AblationHighlight}
        \textbf{(ours) DAHF} {\scriptsize (+ Directional scoring)}
        & \ablationcell{\textbf{87.22}}{+0.93} & \ablationcell{\textbf{77.51}}{+0.79} & \ablationcell{\textbf{89.75}}{+0.38} & \ablationcell{\textbf{85.41}}{+0.27} & \ablationcell{\textbf{87.65}}{+0.20} & \ablationcell{\textbf{77.84}}{+1.08} & \ablationcell{\textbf{35.23}}{+0.69} & \ablationcell{\textbf{41.31}}{+0.99} & \ablationcell{\textbf{39.92}}{+1.44}
        & \ablationcell{\textbf{69.09}}{+0.75} \\
        \bottomrule
    \end{tabular}}
    \arrayrulecolor{black}
\end{table*}

\paragraph{Grouping and direction.}
Edge-based layerwise grouping gains \(0.21\) mean accuracy points over Paris-PEGFAN. Adding directional scoring gains a further \(0.54\) accuracy points and \(0.43\) Macro-F1 points (Table~\ref{tab:structure-accuracy}; Appendix~\ref{app:structure-only-ablation}). Both metrics improve on CiteSeer, Actor, and Chameleon-filtered. PubMed is nearly unchanged, while Macro-F1 decreases on Wisconsin and Cornell.

\paragraph{Direction versus variation magnitude.}
Both scoring rules use node features, but the control maximizes feature variation rather than directional concentration. DAHF gains \(0.48\) mean accuracy points and \(0.79\) Macro-F1 points (Appendix~\ref{app:variance-control}). Accuracy improves on seven datasets and ties on two, while Macro-F1 decreases on Wisconsin and Actor.

\paragraph{Haar support.}
Compared with Paris-PEGFAN, matched-size impulse-response supports have lower mean density and longer mean graph distances on eight of nine datasets (Appendix~\ref{app:support-audit}). Texas shows the opposite pattern.

\section{Conclusion and Limitations}
DAHF adapts the partition tree so that the same Haar transform compares children whose features vary along local directions at different locations and scales. In the periodic-grid experiment, direction-adaptive grouping captures more cross-stripe contrasts in the finest-scale high-pass band. Across nine node classification benchmarks, direction-adaptive grouping improves the average classification scores over edge-based layerwise grouping, with gains varying by dataset.

Our experiments use a transductive setting in which tree construction has access to all graph edges and node features. Generalization to unseen graphs remains to be evaluated. The permutation-equivariance result assumes that the nodes and the given partition tree are relabeled together.

\section*{Impact Statement}
This work studies multiscale representations of graph node features to help classifiers use local contrasts. The experiments use synthetic signals and public graph benchmarks. When applied to graphs involving people, edges and node features may encode existing biases or sensitive information. Such applications would require assessment of privacy risks and errors across groups.

\section*{AI Disclosure Statement}
We used generative AI tools to help identify and organize related literature, discuss experimental controls, write and debug portions of experimental code, summarize results, and draft, translate, and revise the manuscript, figures, tables, and pseudocode. Reported numerical results were checked against experiment outputs or records, and citations were checked against primary papers or publication records. The authors take responsibility for the final submission and its scientific conclusions.

\expandafter\def\expandafter\UrlBreaks\expandafter{\UrlBreaks
  \do\0\do\1\do\2\do\3\do\4\do\5\do\6\do\7\do\8\do\9
  \do\a\do\b\do\c\do\d\do\e\do\f}
\bibliographystyle{icml2027}
\bibliography{references}

\onecolumn
\raggedbottom
\appendix
\section{Related Work}
\label{sec:related-work}
\label{app:related-work}

\subsection{Graph Wavelets and Framelets}
There are two common approaches to constructing graph wavelets and framelets. Spectral methods use the spectrum of the graph Laplacian to construct multiscale filters \citep{hammond2011wavelets,dong2017sparse}. Diffusion wavelets use powers of a diffusion operator to construct a multiresolution representation \citep{coifman2006diffusion}. Spatial methods build multiscale representations directly on the graph vertices. For example, \citet{gavish2010multiscale} construct Haar-like orthonormal bases from nested partitions and relate the decay of wavelet coefficients to smoothness under a tree-induced metric. Given a partition tree, PEGFAN computes a normalized sum and all normalized pairwise differences between the scaling vectors of children sharing a parent \citep{li2024permutation}. The resulting low-pass and high-pass components separate coarse trends from local variation while preserving total energy. The coarsest low-pass and the high-pass coefficients at all scales suffice for exact reconstruction. Lifting-based graph wavelets can also adapt the transform to graph structure and node features \citep{xu2022lifting}. PEGFAN combines multiscale responses with propagation features using a classifier adapted from FSGNN \citep{li2024permutation,maurya2021improving}. DAHF supplies a direction-adaptive partition tree to this pipeline. Beyond graphs, wavelet-based tokenization also separates coarse and fine structures for time series foundation models \citep{masserano2025wavelet}.

\subsection{Directional Wavelets and Framelets}
Euclidean directional wavelets use angularly selective filters to analyze oriented structures at multiple scales. For example, steerable systems combine responses from a fixed bank of directional filters to synthesize a response at a desired orientation \citep{freeman1991steerable,unser2013unifying}. \citet{xiao2021adaptive} construct adaptive directional Haar tight framelets from hierarchical partitions on bounded Euclidean domains. For directed graphs, they derive two undirected graphs from \(WW^\top\) and \(W^\top W\), where \(W\) is the adjacency matrix. They obtain a separate hierarchy for each graph, then use the Cartesian product of the two interval partitions to represent vertices as blocks in the unit square. The geometry of these blocks defines directions. In DAHF, directions come from node feature variation on an undirected graph. Feature axes guide which children share a parent and thereby determine the support of the complete Haar framelet system on the graph.

\subsection{Hierarchical Clustering for Multiscale Analysis}
Hierarchical clustering determines which children share a parent and enter the same local Haar transform. The original PEGFAN experiments build a partition tree by applying Ward clustering \citep{ward1963hierarchical} and balanced cuts to adjacency structure at successive levels \citep{li2024permutation}. Paris defines a distance by sampling node pairs and uses it to merge clusters progressively into a hierarchical clustering tree \citep{bonald2018hierarchical}. The graph framelet example code supplied by Jianfei Li includes Paris clustering and balanced cuts.\footnote{The supplied \texttt{Graph-Framelet-Transform} snapshot is pinned to commit \texttt{13a38cd11036412472c2470639fd2764546f3e11}; see \url{https://github.com/JFLi11/Graph-Framelet-Transform}. Its \texttt{main\_demo.py} calls \texttt{Paris()} and \texttt{cut\_balanced}. This example is separate from the original PEGFAN implementation.} In our preliminary experiments, Paris and Ward gave similar node classification results. We therefore use the example's Paris clustering procedure to construct the Paris-PEGFAN reference. Feature geometry can also guide hierarchy construction. \citet{ram2011generalized} order samples by feature distances at each level so that wavelet filtering acts on a sequence expected to vary more smoothly. DAHF first fits a set of candidate axes to feature differences between adjacent nodes and shares these axes across levels. At each grouping step, DAHF filters candidates using the reciprocal of a Paris-style distance and uses the axes to score the proposed parents that pass the filtering. For each proposed parent, we find the axis along which feature variation among the children is most concentrated and use the proportion of total variation along that axis as the directional score. We weight feature variation by the number of original nodes in each child. Feature variation can thus affect the local neighborhoods used for Haar analysis at each scale.

\definecolor{PedafTablePink}{HTML}{E5B5B5}
\definecolor{PedafTableTeal}{HTML}{C5E9E3}
\definecolor{PedafTableBlue}{HTML}{AFC3D8}
\newenvironment{pedafappendixtable}{%
    \begingroup
    \ifdefined\pedafTableSerif\pedafTableSerif\else\rmfamily\fi
    \fontsize{9.3}{11.2}\selectfont
    \setlength{\tabcolsep}{4.2pt}
    \renewcommand{\arraystretch}{1.12}
    \arrayrulecolor{black}
}{\endgroup}
\newcommand{\pedafTableMark}[2]{%
    \begingroup\setlength{\fboxsep}{1pt}%
    \colorbox{#1}{\strut #2}\endgroup}

\section{Fitting the Direction Dictionary}
\label{app:direction-dictionary}

\paragraph{Sampling edges and forming samples.}
Let \(M\) be the number of distinct undirected edges after removing self-loops. With seed \(42\), sample \(\min(M,8192)\) edges uniformly without replacement. Each edge \((u,v)\) yields a feature difference \(x_u-x_v\). Discard differences with norm at most \(10^{-12}\) and normalize the remaining differences to unit vectors \(z_k\). If no samples remain, all proposed parents receive a directional score of zero.

\paragraph{Initializing candidate axes.}
Choose the first axis \(q_1\) uniformly from the retained samples. Given \(r\) axes, compute each sample's largest absolute inner product with them. Use the sample with the smallest such value as the next axis:
\[
    k^\star\in\operatorname*{arg\,min}_{k}
        \max_{1\leq j\leq r}|z_k^\top q_j|,
    \qquad
    q_{r+1}=z_{k^\star}.
\]
Stop when the dictionary reaches \(R\) axes or when the smallest of these maximum absolute inner products reaches \(1-10^{-8}\). In the latter case, every retained sample already aligns closely with one of the axes.

\paragraph{Assigning samples and updating axes.}
Run ten assignment and update rounds. In each round, let \(r(k)\) index the axis with the largest absolute inner product with \(z_k\). For each axis \(q_r\), align the signs of its assigned samples, sum them, and normalize the sum:
\[
    b_r=\sum_{k:\,r(k)=r}\operatorname{sgn}(z_k^\top q_r)z_k,
    \qquad
    q_r^{\mathrm{new}}=\frac{b_r}{\|b_r\|_2},
\]
Here \(\operatorname{sgn}(0)=1\). If an axis has assigned samples but \(\|b_r\|_2\leq10^{-12}\), keep that axis for the current round to avoid amplifying numerical error during normalization.

\paragraph{Handling an empty assignment.}
Using the \(K\) axes at the start of the round, compute each sample's best alignment:
\[
    a_k=\max_{1\leq j\leq K}|z_k^\top q_j|.
\]
Replace the empty axis with the sample that minimizes \(a_k\). Break ties by taking the first sample in the retained order.

\section{Analysis of Tree Construction Ordering}
\label{app:ordering}

\subsection{Tree Construction Pseudocode}
\label{app:tree-algorithm}

\begin{algorithm}[H]
    \caption{Construct a Direction-Adaptive Partition Tree}
    \label{alg:direction-tree}
    \footnotesize
    \begin{tabbing}
        \hspace{1em}\=\hspace{1em}\=\hspace{1em}\=\kill
        \textbf{Input}\quad Loop-free adjacency \(A\), normalized features \(X\) \\
        \> direction limit \(R\), child limit \(h\), threshold \(\tau\) \\
        Fit candidate axes \(Q\) on the original graph (Appendix~\ref{app:direction-dictionary}). \\
        \(\ell\gets0,\ A^{(0)}\gets A\), and let \(V^{(0)}\) be the original nodes. \\
        Initialize tree \(\mathcal T\) with \(V^{(0)}\). \\
        \textbf{while} \(\lvert V^{(\ell)}\rvert>h\) and \(A^{(\ell)}\) has a positive-weight edge \\
        \> \(B\gets V^{(\ell)}\), \(\mathcal P\gets\emptyset\) (unassigned vertices and groups) \\
        \> \textbf{while} a positive-weight edge has both endpoints in \(B\) \\
        \> \> Retain these edges using \(P_\ell(\{i\},j)\geq\tau P_{\max}\). \\
        \> \> Choose retained edge \(\{i,j\}\) maximizing \(C(\{i,j\})\). \\
        \> \> \(S\gets\{i,j\}\), \(B\gets B\setminus S\) \\
        \> \> \textbf{while} \(\lvert S\rvert<h\) and \(\mathcal J(S)\ne\emptyset\) \\
        \> \> \> Retain \(j\in\mathcal J(S)\) using \(P_\ell(S,j)\geq\tau P_{\max}\). \\
        \> \> \> Choose retained \(j\) maximizing \(C(S\cup\{j\})\). \\
        \> \> \> \(S\gets S\cup\{j\}\), \(B\gets B\setminus\{j\}\) \\
        \> \> \(\mathcal P\gets\mathcal P\cup\{S\}\) \\
        \> \(\mathcal P\gets\mathcal P\cup\{\{v\}:v\in B\}\) (retain singleton groups) \\
        \> Link each group's children to its parent to form level \(\ell+1\). \\
        \> Aggregate edge weights, original node counts, and cluster features. \\
        \> Form \(A^{(\ell+1)}\), \(V^{(\ell+1)}\), and set \(\ell\gets\ell+1\). \\
        \textbf{if} \(\lvert V^{(\ell)}\rvert>1\), attach the remaining vertices to one root. \\
        \textbf{Output}\quad Partition tree \(\mathcal T\)
    \end{tabbing}
\end{algorithm}

\subsection{Paris Merge Heights}

\paragraph{Original Paris distance.}
Let \(v_a\) and \(v_b\) be cluster volumes, the sums of the weighted degrees of their original vertices. Write \(P(a,b)\) for the reciprocal of the Paris distance. For three disjoint clusters \(a,b,c\), additivity of volumes and intercluster edge weights gives
\[
    P(a\cup b,c)
    =\frac{v_a}{v_a+v_b}P(a,c)+\frac{v_b}{v_a+v_b}P(b,c).
\]
The new similarity therefore does not exceed the larger of \(P(a,c)\) and \(P(b,c)\) \citep{bonald2018hierarchical}. If \(a,b\) are the most similar current pair, all new similarities are at most \(P(a,b)\). Taking reciprocal similarities as merge heights, repeatedly merging the closest pair gives nondecreasing merge heights.

\subsection{A Counterexample to Direct Directional Weighting}
\label{app:paris-angle-counterexample}
Consider a triangle whose three edge weights are all \(1\). Nodes \(a,b,c\) have feature coordinates \((-1,-1/3)\), \((1,1/3)\), and \((0,1)\), respectively. Use the horizontal and vertical axes as fixed candidate directions. Let \(P(a,b)\) be the reciprocal Paris distance and \(D(a,b)\) the maximum squared cosine between the difference of two cluster means and a candidate axis. Introduce a direction weight by \(P_\lambda=P(1+\lambda D)\), where \(\lambda>0\). All three original pairwise similarities equal \(3/2\), and their direction scores are
\[
    D(a,b)=\frac{9}{10},\quad
    D(a,c)=\frac{16}{25},\quad
    D(b,c)=\frac{9}{13}.
\]
Thus, \(a,b\) merge first. Their merged mean is \((0,0)\), whose difference from \(c\) is exactly vertical, raising the direction score to \(1\). The structural similarity remains \(3/2\), by the update formula in Appendix~\ref{app:ordering}, so
\[
    P_\lambda(a\cup b,c)
    =\tfrac32(1+\lambda)>P_\lambda(a,b).
\]
Since merge heights are reciprocal similarities, the second merge is lower than the first.

\subsection{Proof of Theorem~\ref{thm:candidate-order}}

\begin{proof}
\textbf{(1) Uniqueness of candidate selection.}
Fix a seed-selection or parent-growth state. Let \(\mathcal C\) be its finite nonempty candidate set and \(P(e)>0\) the structural score of candidate \(e\). Define
\[
    P_{\max}=\max_{e\in\mathcal C}P(e),\qquad
    \mathcal C_\tau=\{e\in\mathcal C:P(e)\geq\tau P_{\max}\}.
\]
Since \(0\leq\tau\leq1\),
\[
    \emptyset\ne\operatorname*{arg\,max}_{e\in\mathcal C}P(e)
    \subseteq\mathcal C_\tau.
\]
For the structurally retained candidates, apply the following rules:

\textbf{Seed selection.} Prefer the edge with the larger directional score \(C(\{i,j\})\). If tied, compare structural scores, then edge weights, preferring larger values. If these also tie, choose the smaller endpoint indices, comparing \(i\) before \(j\), with \(i<j\).

\textbf{Parent growth.} Prefer the candidate with the larger directional score \(C(S\cup\{j\})\). If tied, prefer the larger structural score; if still tied, choose the smaller candidate index \(j\).

In both rules, a later criterion is compared only when all earlier criteria tie. Distinct seed edges have different ordered endpoint pairs, and distinct growth candidates have different indices. Thus, every pair of distinct candidates is distinguished by these comparisons. Each criterion has a consistent order, so these rules define a strict total order on the finite nonempty set \(\mathcal C_\tau\), with exactly one preferred candidate.

Let \(e^\star\) be the unique preferred candidate. Membership in \(\mathcal C_\tau\) gives \(P(e^\star)\geq\tau P_{\max}\). If \(\tau=1\) and the structural maximum is unique, then \(\mathcal C_1=\{e^\star\}\), so this step depends only on the structural score.

\textbf{(2) Nested partitions and finite termination.}
Let \(V\) be the original node set and \(\mathcal I_\ell\) the level-\(\ell\) cluster index set, with partitions
\[
    \mathcal P_\ell=\{V_i^{(\ell)}:i\in\mathcal I_\ell\},\qquad
    \mathcal P_0=\{\{u\}:u\in V\}.
\]
Selected children are immediately marked assigned, and unselected remaining clusters are retained as singleton groups. Thus, every child has exactly one parent, defining a surjection
\[
    \pi_\ell:\mathcal I_\ell\longrightarrow\mathcal I_{\ell+1},\qquad
    \mathcal I_\ell=\bigsqcup_{a\in\mathcal I_{\ell+1}}\pi_\ell^{-1}(a).
\]
The next-level parents satisfy
\[
    V_a^{(\ell+1)}=\bigsqcup_{i\in\pi_\ell^{-1}(a)}V_i^{(\ell)},\qquad
    \bigsqcup_{a\in\mathcal I_{\ell+1}}V_a^{(\ell+1)}=V.
\]
Induction from \(\mathcal P_0\) shows that every level is a partition and \(\mathcal P_\ell\preceq\mathcal P_{\ell+1}\), where \(\preceq\) means that the left partition refines the right partition.

Write \(n_\ell=|\mathcal I_\ell|\). Each regular round starts with a positive-weight edge. Part (1) ensures that at least one seed pair is selected, so at least one parent contains two children. Hence
\[
    n_\ell-n_{\ell+1}
    =\sum_{a\in\mathcal I_{\ell+1}}\bigl(|\pi_\ell^{-1}(a)|-1\bigr)
    \geq1.
\]
The positive integer cluster count ensures that the regular rounds terminate. If multiple clusters remain, they are attached to a single root. Finally, \(\mathcal I_L=\{r\}\), \(n_L=1\), and \(L\leq n_0-1\).

\textbf{(3) Strict hierarchical ordering of the entire tree.}
Represent each tree vertex by \((\ell,i)\) to distinguish clusters at different levels. Define the parent--child edge set and height function by
\[
    E_{\mathcal T}
    =\bigl\{((\ell,i),(\ell+1,\pi_\ell(i))):0\leq\ell<L,\ i\in\mathcal I_\ell\bigr\},
    \qquad H(\ell,i)=\ell.
\]
Every edge directed from a child to its parent satisfies
\[
    H(\ell+1,\pi_\ell(i))-H(\ell,i)=1>0.
\]
If there were a directed cycle \(x_0\to x_1\to\cdots\to x_q=x_0\) with \(q\geq1\) edges, then
\[
    0=H(x_q)-H(x_0)
    =\sum_{t=0}^{q-1}\bigl(H(x_{t+1})-H(x_t)\bigr)=q>0,
\]
a contradiction. Every nonroot vertex has exactly one parent, and following parents for \(L-\ell\) steps reaches the unique \((L,r)\). Thus, the output is a uniquely rooted tree whose heights strictly increase along every leaf-to-root path.
\end{proof}

\section{Wavelet Background and Recursive Haar Operators}
\label{app:haar-operators}

\subsection{Wavelet and Haar Background}
\label{app:wavelet-background}

\paragraph{Position and scale.}
A wavelet measures how a signal varies within a local region. Starting from a mother wavelet \(\psi\), a scaled and translated copy is
\[
    \psi_{s,t}(u)=s^{-1/2}\psi\!\left(\frac{u-t}{s}\right),\qquad s>0.
\]
The translation \(t\) sets the position and the scale \(s\) controls the width. A coefficient \(\langle f,\psi_{s,t}\rangle\) measures how strongly the local signal \(f\) matches this pattern. Smaller scales capture changes over shorter intervals, and larger scales capture changes over longer intervals. A companion scaling function \(\phi\) measures the coarse signal level \citep{mallat1989multiresolution}.

\paragraph{The Haar wavelet.}
The Haar scaling function is \(\phi(u)=\mathbf{1}_{[0,1)}(u)\), where \(\mathbf{1}_I\) is the indicator of interval \(I\). The Haar mother wavelet is
\[
    \psi(u)=\mathbf{1}_{[0,1/2)}(u)-\mathbf{1}_{[1/2,1)}(u).
\]
Its positive and negative halves cancel for a constant signal. When the two halves have different averages, the coefficient gives a signed measure of that difference. For adjacent samples \(y_1,y_2\), the discrete Haar step is
\[
    a=\frac{y_1+y_2}{\sqrt{2}},
    \qquad
    d=\frac{y_1-y_2}{\sqrt{2}}.
\]
The low-pass coefficient \(a\) is a normalized sum and the high-pass coefficient \(d\) is a normalized difference. Together they recover both samples via \(y_1=(a+d)/\sqrt{2}\) and \(y_2=(a-d)/\sqrt{2}\). They also preserve energy, \(y_1^2+y_2^2=a^2+d^2\).

\paragraph{From pairs to multiple scales.}
In the discrete Haar transform, we retain the high-pass coefficients and apply the same pairwise step to the low-pass coefficients. For example, for the signal \((3,1,0,0)\), the first step gives low-pass coefficients \((2\sqrt{2},0)\) and high-pass coefficients \((\sqrt{2},0)\). The next step gives a low-pass coefficient \(2\) and a high-pass coefficient \(2\). The nonzero high-pass coefficient at the first scale records the difference within the first pair, while the one at the second scale records the difference between pairs. The coarsest low-pass coefficient and the high-pass coefficients at both scales suffice to recover the four samples.

\paragraph{From signals on a line to signals on a graph.}
On a line, consecutive pairs determine the local Haar comparisons at each scale. On a graph, the children sharing a parent in the partition tree determine those comparisons. A parent with two children uses exactly the Haar step above. For a parent with \(m\) children, PEGFAN uses a normalized sum and records all \(\binom{m}{2}\) normalized pairwise differences \citep{li2024permutation}. For \(m>2\), this produces more coefficients than child signals. This redundant representation still preserves energy and permits exact reconstruction, as described in Appendix~\ref{app:haar-parseval}.

\subsection{Proof of Proposition~\ref{prop:wavelet-rotation}}
\label{app:wavelet-rotation-proof}

\begin{proof}
Set \(v=R_\theta^{-1}u\), so \(u=R_\theta v\). A rotation has determinant \(1\), hence \(du=dv\). Substitution in the inner product gives
\begin{align*}
    \langle f_b,\psi_\theta\rangle
    &=\int_{\mathbb R^2}f_b(u)\,
      \overline{\psi(R_\theta^{-1}u)}\,du\\
    &=\int_{\mathbb R^2}f_b(R_\theta v)\,
      \overline{\psi(v)}\,dv
      =\langle f_b\circ R_\theta,\psi\rangle.
\end{align*}
\end{proof}

\subsection{Haar Operations at a Parent}

Consider a parent \(V\) with \(m\) children. If child \(i\) is a leaf, \(y_i\) is its node's input feature vector in \(Z\). Otherwise, \(y_i\) is the low-pass signal already computed at that child. Stack the \(y_i^\top\) as the rows of \(Y\). Following PEGFAN \citep[Corollary~1]{li2024permutation}, define \(\boldsymbol p_V=\mathbf1/\sqrt m\) and let the rows of \(\boldsymbol B_V\) be \((e_i-e_j)^\top/\sqrt m\) for \(i<j\), where \(e_i\) is the \(i\)th coordinate unit vector. The parent's low-pass and high-pass coefficients are
\[
    Y_{\mathrm{low}}=\boldsymbol p_V^\top Y,\qquad
    Y_{\mathrm{high}}=\boldsymbol B_VY.
\]
The diagonal entries of \(\boldsymbol B_V^\top\boldsymbol B_V\) are \((m-1)/m\), and all off-diagonal entries are \(-1/m\). Adding \(\boldsymbol p_V\boldsymbol p_V^\top\) gives the identity. Thus, the operations at one parent satisfy
\[
    \|Y_{\mathrm{low}}\|_F^2+\|Y_{\mathrm{high}}\|_F^2=\|Y\|_F^2,\qquad
    Y=\boldsymbol p_VY_{\mathrm{low}}+\boldsymbol B_V^\top Y_{\mathrm{high}}.
\]

\subsection{Recursive Operators}

Let \(n\) be the number of nodes in the original graph. At level \(0\), set \(\Phi_0=I_n\), so multiplying row \(u\) of \(\Phi_0\) by \(Z\) extracts the feature vector of node \(u\).

Suppose a parent at level \(\ell+1\) has \(m\) children, with row \(\phi_i^\top\) of \(\Phi_\ell\) corresponding to child \(i\). The normalized sum of these rows yields the parent's low-pass row. The normalized difference of two rows yields a high-pass row:
\[
    \underbrace{\frac{1}{\sqrt m}\sum_{i=1}^{m}\phi_i^\top}_{\text{one row of }\Phi_{\ell+1}},
    \qquad
    \underbrace{\frac{\phi_i^\top-\phi_j^\top}{\sqrt m}}_{\text{one row of }W_\ell}
    \quad (1\leq i<j\leq m).
\]
Stack the sum rows from all parents at level \(\ell+1\) to form \(\Phi_{\ell+1}\) and their difference rows to form \(W_\ell\). Repeat until obtaining the root low-pass operator \(\Phi_L\) and the high-pass operators \(W_0,\ldots,W_{L-1}\).

\subsection{Parseval Tight Frame}
\label{app:haar-parseval}

Stack the root low-pass operator \(\Phi=\Phi_L\) and the high-pass operators \(W_\ell\) to obtain the analysis matrix \(F\). For a scalar node signal \(z\in\mathbb R^n\), \(Fz\) contains all retained coefficients. The rows of \(F\) form a Parseval tight frame for \(\mathbb R^n\) if \(F^\top F=I_n\). This identity preserves energy, \(\|Fz\|_2^2=\|z\|_2^2\), and lets the adjoint reconstruct the signal exactly, \(F^\top(Fz)=z\).

Applying the local energy identity at all parents of level \(\ell+1\) shows that the level-\(\ell\) low-pass energy equals the next-level low-pass energy plus the high-pass energy produced by this transition. Since this holds for every \(z\), we have
\[
    \Phi_\ell^\top\Phi_\ell
    =\Phi_{\ell+1}^\top\Phi_{\ell+1}+W_\ell^\top W_\ell.
\]
Starting at \(\Phi_0=I_n\) and substituting recursively gives
\[
    F^\top F=\Phi^\top\Phi+\sum_{\ell=0}^{L-1}W_\ell^\top W_\ell=I_n.
\]
For a multichannel input \(Z\), these identities apply column by column.

\subsection{Sparsity}
\label{app:haar-sparsity}

We adapt PEGFAN's nonzero-coefficient counting argument \citep[Theorem~2]{li2024permutation} to level-dependent child limits and explicitly count the root low-pass coefficient. The operator recursion shows that each low-pass row is nonzero only on the original nodes of its cluster. A high-pass row is the difference of two such low-pass rows, so its support lies within that pair of children.

An original node \(u\) belongs to exactly one child at level \(\ell\). If its parent has \(m\) children, at most \(m-1\) high-pass coefficients at that level involve \(u\), one for each difference between its child and another child. Thus, \(z_u\) affects at most \(h_\ell-1\) high-pass coefficients at that level and one root low-pass coefficient. Every nonzero coefficient involves at least one nonzero input entry. Counting and summing over these entries gives the bound in Theorem~\ref{thm:haar-sparsity}. Using the actual maximum child count at each level also covers the final added root.

\subsection{Permutation Equivariance Given the Partition Tree}

Following PEGFAN \citep[Theorem~4 and Proposition~2]{li2024permutation}, fix the partition tree and relabel nodes together with their corresponding cluster memberships. Let \(\Pi\) be the node permutation matrix, so the relabeled input is \(\Pi Z\). Reordering children of a parent can reorder high-pass coefficients and reverse some pairwise differences. A signed permutation matrix \(U_\ell\) represents these changes, giving \(\widetilde W_\ell=U_\ell W_\ell\Pi^\top\). Since \(U_\ell^\top U_\ell=I\), mapping back to nodes yields
\begin{equation}
    \widetilde W_\ell^\top\widetilde W_\ell\,\Pi Z
    =\Pi W_\ell^\top W_\ell Z
    =\Pi H_\ell.
    \label{eq:haar-equivariance}
\end{equation}
The root low-pass response follows the same node permutation. Thus, given the partition tree, Haar node responses at all scales are relabeled along with the nodes.

\section{Per-Dataset Results for DAHF and Preset Paris-PEGFAN}

\begin{table}[H]
    \centering

    \caption{Mean test scores (\%) for preset Paris-PEGFAN and DAHF on each dataset. On Cornell, the child limits are \(8\) and \(4\), respectively. Pink marks higher scores; teal marks the average row.}
    \label{tab:pedaf-paris}
    \begin{pedafappendixtable}
\begin{tabular}{lrrrr}
        \toprule
\rowcolor{white}
        & \multicolumn{2}{c}{Accuracy (\%)} & \multicolumn{2}{c}{Macro-F1 (\%)} \\
        \cmidrule(lr){2-3}\cmidrule(lr){4-5}
\rowcolor{white}
\textbf{Dataset} & \textbf{Paris-PEGFAN} & \textbf{DAHF} & \textbf{Paris-PEGFAN} & \textbf{DAHF} \\
        \midrule
\rowcolor{white}
Cora & 86.30 & \pedafTableMark{PedafTablePink}{87.22} & 85.14 & \pedafTableMark{PedafTablePink}{85.96} \\
\rowcolor{black!2}
CiteSeer & 76.72 & \pedafTableMark{PedafTablePink}{77.51} & 72.17 & \pedafTableMark{PedafTablePink}{73.10} \\
\rowcolor{white}
PubMed & 89.37 & \pedafTableMark{PedafTablePink}{89.75} & 88.88 & \pedafTableMark{PedafTablePink}{89.39} \\
\rowcolor{black!2}
Texas & 85.14 & \pedafTableMark{PedafTablePink}{85.41} & 59.36 & \pedafTableMark{PedafTablePink}{60.88} \\
\rowcolor{white}
Wisconsin & 87.45 & \pedafTableMark{PedafTablePink}{87.65} & 69.16 & \pedafTableMark{PedafTablePink}{69.71} \\
\rowcolor{black!2}
Cornell & 76.76 & \pedafTableMark{PedafTablePink}{77.84} & 55.51 & \pedafTableMark{PedafTablePink}{60.31} \\
\rowcolor{white}
Actor & 34.54 & \pedafTableMark{PedafTablePink}{35.23} & 24.32 & \pedafTableMark{PedafTablePink}{27.06} \\
\rowcolor{black!2}
Chameleon-filtered & 40.31 & \pedafTableMark{PedafTablePink}{41.31} & 33.81 & \pedafTableMark{PedafTablePink}{34.36} \\
\rowcolor{white}
Squirrel-filtered & 38.48 & \pedafTableMark{PedafTablePink}{39.92} & 25.83 & \pedafTableMark{PedafTablePink}{27.41} \\
        \midrule
\rowcolor{PedafTableTeal}
\textbf{Avg.} & 68.34 & \pedafTableMark{PedafTablePink}{69.09} & 57.13 & \pedafTableMark{PedafTablePink}{58.69} \\
        \bottomrule
    
\end{tabular}
\end{pedafappendixtable}
\end{table}

\section{Training and Model Selection}
\label{app:pedaf-selection}

\paragraph{Data and metrics.}
Cora, CiteSeer, PubMed, Texas, Wisconsin, Cornell, and Actor use the ten fixed Geom-GCN splits \citep{pei2020geom}. Chameleon and Squirrel use the filtered versions with duplicate nodes removed and their ten provided splits \citep{platonov2023critical}. All models share graph inputs and training, validation, and test splits. Training uses all graph edges and node features, with supervision restricted to training nodes.

Test accuracy and Macro-F1 are computed on each split and then averaged over ten splits. Macro-F1 weights the classwise F1 scores equally, and the overall mean weights datasets equally. Sample standard deviations use the \(n-1\) denominator, with \(n=10\). For propagation, DAHF and Paris-PEGFAN include self-loops on Cora, CiteSeer, and PubMed and omit them on the remaining datasets.

\paragraph{Propagation convention.}
We follow the normalized propagation in the corrigendum in PEGFAN's official repository.\footnote{See the corrigendum in \url{https://github.com/zrgcityu/PEGFAN}.} With the dataset-specific self-loop choice above, write \(A_\ast=A+I\) for the citation graphs and \(A_\ast=A\) otherwise, and let \(D_\ast\) be its degree matrix. Our notation is \(\widehat A=D_\ast^{-1/2}A_\ast D_\ast^{-1/2}\), with zero inverse degree for isolated nodes. Type~b applies Haar to \(X\), and Type~c applies Haar to \(\widehat A X\), following PEGFAN's channel arrangement. These input settings and the adjacency-based tree used in the original paper are distinct from our choice of Paris clustering for the reference implementation.

\paragraph{Training.}
Each split uses seed \(42\) and trains for at most \(1500\) epochs. Training stops after \(100\) epochs without improvement in validation loss, and we retain the checkpoint with lowest validation loss. For each model and dataset, we select the complete configuration with highest mean validation accuracy over the ten splits and report its test scores on those same splits.

\paragraph{DAHF search grid.}
The main experiment fixes \(R=7\) and evaluates all sixteen combinations of
\[
    \tau\in\{0,0.15,0.30,0.45,0.60,0.75,0.90,1.00\},
    \qquad h\in\{4,8\}.
\]
Table~\ref{tab:pedaf-selection} lists the selected configurations. Each validation or test score averages the ten fixed splits.

\begin{table}[H]
    \centering
    \caption{Configurations and scores for DAHF in the main comparison. Scores are percentages. Teal marks the selected configurations and the average row.}
    \label{tab:pedaf-selection}

    \begin{pedafappendixtable}
\begin{tabular}{@{}lrrrrr@{}}
        \toprule
\rowcolor{white}
\textbf{Dataset} & \textbf{Child limit \(h\)} & \textbf{Threshold \(\tau\)} & \textbf{Val. accuracy} & \textbf{Test accuracy} & \textbf{Test Macro-F1} \\
        \midrule
\rowcolor{white}
Cora & \pedafTableMark{PedafTableTeal}{8} & \pedafTableMark{PedafTableTeal}{0.30} & 87.2487 & 87.2233 & 85.9648 \\
\rowcolor{black!2}
CiteSeer & \pedafTableMark{PedafTableTeal}{8} & \pedafTableMark{PedafTableTeal}{0.00} & 77.7454 & 77.5073 & 73.1023 \\
\rowcolor{white}
PubMed & \pedafTableMark{PedafTableTeal}{8} & \pedafTableMark{PedafTableTeal}{1.00} & 89.9572 & 89.7541 & 89.3910 \\
\rowcolor{black!2}
Texas & \pedafTableMark{PedafTableTeal}{8} & \pedafTableMark{PedafTableTeal}{0.00} & 87.9661 & 85.4054 & 60.8756 \\
\rowcolor{white}
Wisconsin & \pedafTableMark{PedafTableTeal}{8} & \pedafTableMark{PedafTableTeal}{0.90} & 86.7500 & 87.6471 & 69.7074 \\
\rowcolor{black!2}
Cornell & \pedafTableMark{PedafTableTeal}{4} & \pedafTableMark{PedafTableTeal}{0.15} & 78.9831 & 77.8378 & 60.3075 \\
\rowcolor{white}
Actor & \pedafTableMark{PedafTableTeal}{8} & \pedafTableMark{PedafTableTeal}{0.15} & 36.0362 & 35.2303 & 27.0597 \\
\rowcolor{black!2}
Chameleon-filtered & \pedafTableMark{PedafTableTeal}{8} & \pedafTableMark{PedafTableTeal}{0.75} & 42.3259 & 41.3061 & 34.3638 \\
\rowcolor{white}
Squirrel-filtered & \pedafTableMark{PedafTableTeal}{8} & \pedafTableMark{PedafTableTeal}{0.30} & 38.6337 & 39.9151 & 27.4129 \\
        \midrule
\rowcolor{PedafTableTeal}
\textbf{Avg.} &  &  &  & 69.0918 & 58.6872 \\
        \bottomrule
    
\end{tabular}
\end{pedafappendixtable}
\end{table}

\paragraph{Baseline sources.}
The propagation and graph filtering baselines include FSGNN \citep{maurya2021improving}, ACM-GCN and ACM-GCN++ \citep{luan2022revisiting}, GloGNN++ \citep{li2022global}, ChebNetII \citep{he2022chebnetii}, OptBasisGNN \citep{guo2023optimal}, GWN-fa \citep{yue2025graphwave}, GAMMA \citep{ghazizadeh2025gamma}, AdaSpec \citep{liu2026adaspec}, and CASTOR \citep{ghazizadeh2026castor}. SGFormer uses a graph Transformer \citep{wu2023sgformer}.

\paragraph{Baseline search and replay.}
The fourteen tuned baselines, including MLP, GCN \citep{kipf2017semi}, and CPF \citep{li2025partitionwise}, share the same splits and seed. Each tries sixteen configurations per dataset. On CiteSeer, AdaSpec's \texttt{p12} failed twice on one split with a nonfinite gradient norm. We selected \texttt{p10} from the fifteen complete configurations. All \(990\) test replays for the original eleven baselines completed. MLP, GCN, and CPF completed \(4320/4320\) search runs and \(270/270\) test replays, with zero failures in either phase. Paris-PEGFAN uses the preset configuration from our reproduction.

\section{Edge-Based Layerwise Grouping versus Direction-Adaptive Grouping}
\label{app:structure-only-ablation}

\paragraph{Setup.}
This comparison includes preset Paris-PEGFAN, edge-based layerwise grouping, and full DAHF. Full DAHF adds directional scoring to edge-based layerwise grouping. DAHF fixes \(R=7\) and selects one of sixteen configurations combining eight \(\tau\) values with \(h\in\{4,8\}\) for each dataset by mean validation accuracy over ten splits. Each run selects its checkpoint by validation loss. Appendix~\ref{app:pedaf-selection} gives the full selection protocol. Paris-PEGFAN retains its preset configuration. On Cornell, its child limit is \(8\), while both other methods use \(4\). Table~\ref{tab:pedaf-paris} gives per-dataset accuracy and Macro-F1 against Paris-PEGFAN.

Edge-based layerwise grouping selects the highest structural score when choosing seed pairs and growing parents. It resolves ties without node features. The child limit for each dataset is fixed to the value selected in the main experiment. This method and full DAHF both retain the complete Haar channels and use the same classifier design and training settings. We pair runs on the same ten splits, yielding \(90\) pairs across nine datasets. The grouping rule can change the hierarchy and its Haar channels. This comparison measures the resulting effect on the complete classification model.

\begin{table}[H]
    \centering

    \caption{\textbf{Partition-tree ablation with complete Haar channels.} Differences are DAHF minus edge-based layerwise grouping, in percentage points. The value after \(\pm\) is the sample standard deviation of the paired differences over ten splits. Teal marks the average row.}
    \label{tab:structure-ablation}
    \begin{pedafappendixtable}
\begin{tabular}{lrr}
        \toprule
\rowcolor{white}
\textbf{Dataset} & \textbf{\(\Delta\) Accuracy} & \textbf{\(\Delta\) Macro-F1} \\
        \midrule
\rowcolor{white}
Cora & +0.30\,\ensuremath{\pm}\,0.92 & +0.33\,\ensuremath{\pm}\,1.01 \\
\rowcolor{black!2}
CiteSeer & +0.97\,\ensuremath{\pm}\,0.84 & +1.04\,\ensuremath{\pm}\,1.05 \\
\rowcolor{white}
PubMed & \ensuremath{-}0.05\,\ensuremath{\pm}\,0.07 & \ensuremath{-}0.03\,\ensuremath{\pm}\,0.11 \\
\rowcolor{black!2}
Texas & \ensuremath{-}0.27\,\ensuremath{\pm}\,2.69 & +0.07\,\ensuremath{\pm}\,3.70 \\
\rowcolor{white}
Wisconsin & +0.00\,\ensuremath{\pm}\,1.85 & \ensuremath{-}1.64\,\ensuremath{\pm}\,7.95 \\
\rowcolor{black!2}
Cornell & +0.54\,\ensuremath{\pm}\,3.78 & \ensuremath{-}0.98\,\ensuremath{\pm}\,8.45 \\
\rowcolor{white}
Actor & +0.92\,\ensuremath{\pm}\,0.97 & +2.49\,\ensuremath{\pm}\,3.28 \\
\rowcolor{black!2}
Chameleon-filtered & +1.51\,\ensuremath{\pm}\,1.66 & +1.40\,\ensuremath{\pm}\,1.98 \\
\rowcolor{white}
Squirrel-filtered & +0.93\,\ensuremath{\pm}\,1.23 & +1.22\,\ensuremath{\pm}\,1.84 \\
        \midrule
\rowcolor{PedafTableTeal}
\textbf{Avg.} & +0.54 & +0.43 \\
        \bottomrule
    
\end{tabular}
\end{pedafappendixtable}
\end{table}

\paragraph{Per-dataset results.}
Compared with edge-based layerwise grouping, full DAHF improves both accuracy and Macro-F1 on CiteSeer, Actor, and Chameleon-filtered. Both scores are nearly unchanged on PubMed, while Macro-F1 decreases on Wisconsin and Cornell.

\section{Computational Complexity and Runtime}
\label{app:computational-cost}

\subsection{Time and Space Complexity}

Let \(N\) and \(E\) be the numbers of original nodes and undirected edges, \(F\) the feature dimension, and \(n_\ell\) and \(m_\ell\) the numbers of clusters and edges at level \(\ell\), where \(n_0=N\) and \(m_0=E\). Let \(L\) be the root level and \(R\) the dictionary size limit. The bounds below describe the implementation used in the experiments.

\paragraph{Dictionary fitting.}
We sample \(S_0=\min(E,8192)\) edges and retain \(S\le S_0\) nonzero feature differences. With \(I=10\) refinement rounds and \(J\) empty-direction resets in total, fitting takes
\[
O\bigl(NF+N+E+S_0F+(I+1)SFR+JSR\bigr)
\]
time and \(O(NF+N+E+S_0F+SR+RF)\) peak space, including the input. The sample--axis projections dominate each regular round. An empty-direction reset reduces the current projection matrix over all axes, costing \(O(SR)\).

\paragraph{Partition tree.}
At a regular level \(\ell\), let \(s_\ell\) count seed selections. For growth query \(q\), let \(c_{\ell q}\) be the number of boundary candidates and \(e_{\ell q}\) the number remaining after structural screening. With expected constant-time hash operations and linear-time sparse contraction for the one-hot partition, the time per regular level is
\[
O\Bigl(
    N+n_\ell FR+m_\ell\,[F+R+\log(m_\ell+1)]+s_\ell m_\ell
    +\sum_q\bigl[c_{\ell q}\log(c_{\ell q}+1)
    +e_{\ell q}(F+R+\log(e_{\ell q}+1))\bigr]
\Bigr).
\]
Here \(n_\ell FR\) accounts for projecting cluster means, and the \(m_\ell\) terms account for scoring, sorting, and queuing seed edges. Each seed selection negates the structural-score array before binary search, accounting for \(s_\ell m_\ell\). A growth query sorts boundary candidates and evaluates feature scatter only for those that pass the structural filter. The \(N\) term accounts for recording each partition on the original nodes. Sum this cost over the regular levels and add \(O(N+n_\ell(F+R))\) if a final root is added over \(n_\ell\) remaining clusters. Peak space is
\[
O\bigl(N(L+1)+RF+\max_\ell\{m_\ell+n_\ell(F+R)\}\bigr),
\]
including the saved partitions, dictionary, and the buffers for the current sparse level and candidate scores.

\paragraph{Propagation and Haar channels.}
With \(p=3\) propagation hops, sparse propagation takes \(O(p(E+N)F)\) time once the adjacency operators are prepared. Sparse normalization and coalescing add at most \(O((E+N)\log(E+N+1))\). Preparing the parent maps from saved partitions takes at most \(O((L+1)N\log(N+1)+\sum_{\ell=0}^{L}n_\ell\log(n_\ell+1))\) using comparison-sort bounds. Given these maps, low-pass signals are obtained in one pass up the tree. Each high-pass node response is computed from child signals and parent sums, without materializing all pairwise differences or the matrices \(W_\ell\). Lifting the responses back to the original nodes and assembling the cache for \(C=L+p+2\) channels take, together with the Haar analysis,
\[
O\left(F\left[CN+\sum_{\ell=0}^{L}n_\ell
       +\sum_{s=1}^{L}\sum_{j=0}^{s-1}n_j\right]\right)
\]
time. Peak space for this preprocessing is \(O(N(L+1)+E+CNF+F\sum_{\ell=0}^{L}n_\ell)\), including the graph, the cached channels, and the low-pass signals.

\paragraph{Classifier.}
With hidden width \(D\), \(k\) classes, and batch size \(B\), a cached forward pass on \(v\) nodes takes \(O(vCD(F+k)+\lceil v/B\rceil C)\) time. The affine channel projections and the final readout account for the first term, and the channel weights are normalized once per batch. The parameter count is \(P=CD(F+1)+k(CD+1)+C\), and peak batch activation space is \(O(B[C(F+D)+k])\), in addition to \(O(P)\) space for parameters and training state and the fixed feature cache. Cached full-graph inference uses \(v=N\). Backpropagation has the same leading order as the forward pass on the training batches.

\subsection{Measured Runtime}

Table~\ref{tab:runtime-ratios} compares the runtime of DAHF with that of preset Paris-PEGFAN. Per-epoch time includes parameter updates, training and validation evaluation, and checkpoint selection. Full training is measured until early stopping or the epoch limit. Full-graph inference starts from cached channels and includes batch transfers and the forward pass. These measurements exclude dictionary fitting, tree construction, and channel preprocessing.

Across the nine datasets, the median runtime ratios are \(1.20\times\) per epoch, \(1.25\times\) for full training, and \(1.14\times\) for full-graph inference. DAHF is faster on all three measures on Chameleon-filtered and Squirrel-filtered. Cornell has the largest ratios, \(2.17\times\) for full training and \(2.12\times\) for inference. On Cornell, the child limits for DAHF and preset Paris-PEGFAN are \(4\) and \(8\), respectively.

\begin{table}[H]
    \centering
    \caption{Runtime of DAHF divided by the runtime of preset Paris-PEGFAN. A ratio below \(1\) indicates that DAHF is faster.}
    \label{tab:runtime-ratios}

    \begin{pedafappendixtable}
\begin{tabular}{@{}lrrr@{}}
        \toprule
\rowcolor{white}
\textbf{Dataset} & \textbf{Per epoch} & \textbf{Full training} & \textbf{Full-graph inference} \\
        \midrule
\rowcolor{white}
Cora & 1.14\ensuremath{\times} & 1.07\ensuremath{\times} & 1.03\ensuremath{\times} \\
\rowcolor{black!2}
CiteSeer & 1.25\ensuremath{\times} & 1.55\ensuremath{\times} & 1.37\ensuremath{\times} \\
\rowcolor{white}
PubMed & 1.20\ensuremath{\times} & 1.35\ensuremath{\times} & 1.10\ensuremath{\times} \\
\rowcolor{black!2}
Texas & 1.35\ensuremath{\times} & 1.25\ensuremath{\times} & 1.52\ensuremath{\times} \\
\rowcolor{white}
Wisconsin & 1.27\ensuremath{\times} & 1.31\ensuremath{\times} & 1.15\ensuremath{\times} \\
\rowcolor{black!2}
Cornell & 1.74\ensuremath{\times} & 2.17\ensuremath{\times} & 2.12\ensuremath{\times} \\
\rowcolor{white}
Actor & 1.09\ensuremath{\times} & 1.16\ensuremath{\times} & 1.14\ensuremath{\times} \\
\rowcolor{black!2}
Chameleon-filtered & 0.82\ensuremath{\times} & 0.79\ensuremath{\times} & 0.75\ensuremath{\times} \\
\rowcolor{white}
Squirrel-filtered & 0.65\ensuremath{\times} & 0.55\ensuremath{\times} & 0.60\ensuremath{\times} \\
        \bottomrule
    
\end{tabular}
\end{pedafappendixtable}
\end{table}

\section{Hyperparameter Sensitivity}
\label{app:tau-sensitivity}

\begin{figure}[H]
    \centering
    \includegraphics[width=.98\linewidth]{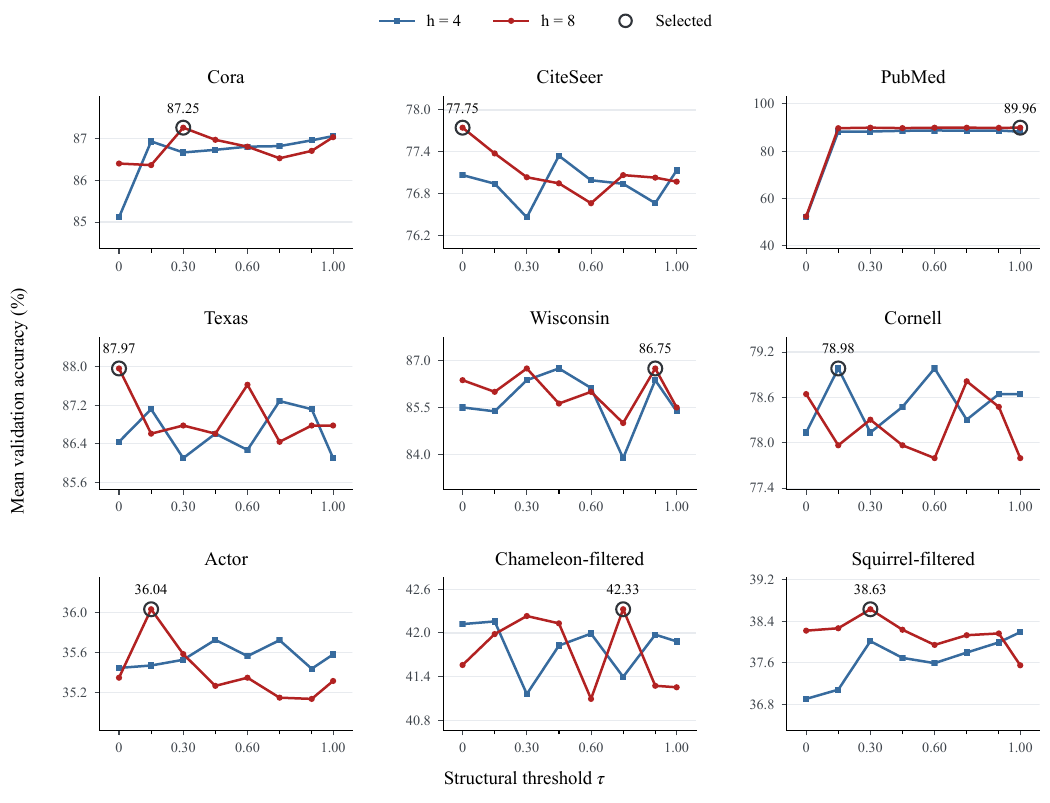}
    \caption{Hyperparameter sensitivity on all nine datasets. All nine panels use a uniform layout and a shared legend. Each point is the mean validation accuracy over ten fixed splits. Open circles mark the configurations selected in the main experiment. Each panel uses its own vertical range.}
    \label{fig:tau-sensitivity}
    \label{fig:tau-sensitivity-appendix}
\end{figure}

We evaluate all eight structural thresholds \(\tau\) at both child limits \(h\in\{4,8\}\), with each score averaging validation accuracy over ten fixed splits. Figure~\ref{fig:tau-sensitivity} gives the complete curves for all nine datasets. Cora and Squirrel-filtered both select \(h=8\) and \(\tau=0.30\), reaching \(87.25\%\) and \(38.63\%\), respectively. At fixed \(h=8\), these scores are \(0.23\) and \(1.08\) percentage points higher than at \(\tau=1\).

\section{Directional Concentration versus Feature Variation Magnitude}
\label{app:variance-control}

\paragraph{Purpose.}
Comparing edge-based layerwise grouping with full DAHF does not distinguish benefits from the magnitude of feature differences from benefits from directional consistency. We therefore compare two tree-building scores that both use node features. The control uses weighted feature variance \(M(U)\), favoring candidates that produce larger feature differences among a parent's children. DAHF uses \(C(U)\), favoring groupings in which these differences concentrate along one candidate direction.

\paragraph{Setup.}
Both variants use the dataset-specific child limit \(h\) and structural threshold \(\tau\) selected in Table~\ref{tab:pedaf-selection}. They share candidate generation, structural screening \(P_\ell\geq\tau P_{\max}\), tie-breaking, and graph coarsening at each level. DAHF fixes \(R=7\) and uses Equation~\eqref{eq:directional-score} to score directional concentration. The feature-variation variant chooses the proposal \(U\) that maximizes the following score among the retained candidates, both when choosing a seed pair and when growing a parent:
\[
    M(U)=
    \frac{\sum_{i\in U}m_i\|\mu_i-\bar\mu_U\|_2^2}{\sum_{i\in U}m_i},
    \qquad
    \bar\mu_U=\frac{\sum_{i\in U}m_i\mu_i}{\sum_{i\in U}m_i}.
\]
Here \(i\) indexes the children in \(U\), \(m_i\) is the number of original nodes in child \(i\), \(\mu_i\) is their mean normalized feature vector, and \(\bar\mu_U\) is the feature mean of the parent weighted by \(m_i\). The feature-variation variant uses \(M(U)\) directly and requires no direction dictionary.

Both variants use the complete Haar transform, the same Type~c propagation, classifier design, and training hyperparameters, and the same ten fixed splits with seed \(42\). Checkpoints are selected by validation loss. We reuse the verified DAHF results from the main comparison and train the feature-variation variant for \(90\) runs. All \(90\) pairs are valid, with no failed runs. The two scoring rules produce different trees on all nine datasets.

\begin{table}[H]
    \centering
    \caption{\textbf{Partition-tree scoring ablation at fixed \((h,\tau)\).}
        Mean accuracy and Macro-F1 are percentages. Each dataset has ten valid pairs.
        \(\Delta\) is DAHF minus the feature-variation variant on each split, reported as mean \(\pm\) sample standard deviation in percentage points.
        The last row weights the nine dataset means equally. Teal marks the average row.}
    \label{tab:variance-control}

    \begin{pedafappendixtable}
\begin{tabular}{@{}lrrcrrc@{}}
        \toprule
\rowcolor{white}
        & \multicolumn{3}{c}{\textbf{Accuracy}} & \multicolumn{3}{c}{\textbf{Macro-F1}} \\
        \cmidrule(lr){2-4}\cmidrule(lr){5-7}
\rowcolor{white}
\textbf{Dataset} & \textbf{DAHF} & \textbf{Variation} & \textbf{\(\Delta\)} & \textbf{DAHF} & \textbf{Variation} & \textbf{\(\Delta\)} \\
        \midrule
\rowcolor{white}
Cora & 87.2233 & 87.0020 & +0.2213\,\ensuremath{\pm}\,0.7847 & 85.9648 & 85.8262 & +0.1386\,\ensuremath{\pm}\,0.7481 \\
\rowcolor{black!2}
CiteSeer & 77.5073 & 76.3297 & +1.1775\,\ensuremath{\pm}\,1.3728 & 73.1023 & 72.0801 & +1.0222\,\ensuremath{\pm}\,1.5387 \\
\rowcolor{white}
PubMed & 89.7541 & 89.4929 & +0.2612\,\ensuremath{\pm}\,0.2397 & 89.3910 & 89.1244 & +0.2666\,\ensuremath{\pm}\,0.2361 \\
\rowcolor{black!2}
Texas & 85.4054 & 85.4054 & 0.0000\,\ensuremath{\pm}\,2.8489 & 60.8756 & 59.9929 & +0.8827\,\ensuremath{\pm}\,5.0574 \\
\rowcolor{white}
Wisconsin & 87.6471 & 87.6471 & 0.0000\,\ensuremath{\pm}\,2.0668 & 69.7074 & 70.3493 & \ensuremath{-}0.6418\,\ensuremath{\pm}\,9.0465 \\
\rowcolor{black!2}
Cornell & 77.8378 & 77.5676 & +0.2703\,\ensuremath{\pm}\,2.9743 & 60.3075 & 57.4668 & +2.8408\,\ensuremath{\pm}\,10.5351 \\
\rowcolor{white}
Actor & 35.2303 & 35.1053 & +0.1250\,\ensuremath{\pm}\,0.6553 & 27.0597 & 27.1135 & \ensuremath{-}0.0538\,\ensuremath{\pm}\,1.3221 \\
\rowcolor{black!2}
Chameleon-filtered & 41.3061 & 40.1350 & +1.1712\,\ensuremath{\pm}\,1.3210 & 34.3638 & 33.1416 & +1.2221\,\ensuremath{\pm}\,1.3734 \\
\rowcolor{white}
Squirrel-filtered & 39.9151 & 38.8145 & +1.1006\,\ensuremath{\pm}\,1.2013 & 27.4129 & 26.0205 & +1.3924\,\ensuremath{\pm}\,2.2243 \\
        \midrule
\rowcolor{PedafTableTeal}
\textbf{Avg.} & 69.0918 & 68.6110 & +0.4808 & 58.6872 & 57.9017 & +0.7855 \\
        \bottomrule
    
\end{tabular}
\end{pedafappendixtable}
\end{table}

\paragraph{Results.}
At fixed \((h,\tau)\), directional concentration improves mean accuracy and Macro-F1 across the nine datasets by \(0.4808\) and \(0.7855\) percentage points, respectively. Accuracy is higher on seven datasets and tied on Texas and Wisconsin. Both metrics improve by more than one percentage point on CiteSeer, Chameleon-filtered, and Squirrel-filtered. Macro-F1 is lower on Wisconsin and Actor. Cornell has the largest mean Macro-F1 gain, \(2.8408\) points, and the largest sample standard deviation of the paired differences across splits, \(10.5351\) points. These comparisons describe observed means. Statistical significance requires a separate test.

\section{Synthetic Stripe Experiment}
\label{app:grid-experiment}

In Figure~\ref{fig:grid-experiment}, Ground truth is the original \(\pm1\) stripe signal. The other columns are Paris-PEGFAN, edge-based layerwise grouping, and full DAHF, respectively. The colorbar applies only to their Haar responses, with red for positive, blue for negative, and white for zero. Heatmaps share one node relabeling, and percentages average finest-scale energy shares over ten relabelings.

We assign values \(+1\) and \(-1\) to alternating columns or rows of the same \(16\times16\) periodic grid to obtain two stripe signals. Every node has a neighbor with the opposite value, but the finest-scale Haar transform compares nodes that share a parent. We compare Paris-PEGFAN, edge-based layerwise grouping, and full DAHF. All three use the same scalar signal after one propagation step and the same Haar transform, with at most two children per ordinary parent. Full DAHF uses \(\tau=0.5\) and fits the direction dictionary and builds the tree from the two-dimensional node features \((0.5+0.25 f_u,\,0.5-0.25 f_u)\), where \(f_u\) is the stripe value. For each pattern, we build trees under ten node relabelings, sharing the relabeling across the three methods in each replicate.

We measure the fraction of the total energy of the signal fed into the Haar transform that enters the finest-scale high-pass band. White cells in the response panels of Figure~\ref{fig:grid-experiment} denote zero finest-scale responses. Nodes paired along the same stripe yield a zero difference. The direction-adaptive tree places more cross-stripe neighbors in the same parent, producing nonzero responses at more nodes.

For the column and row patterns, Paris-PEGFAN captures \(43.4\%\) and \(46.7\%\) of the input energy in the finest-scale high-pass band. Edge-based layerwise grouping yields the same shares. Full DAHF increases them to \(85.1\%\) and \(86.2\%\), and the shares of zero-response nodes fall from \(56.6\%\) and \(53.3\%\) to \(14.9\%\) and \(13.8\%\), respectively. On this grid, regrouping brings more cross-stripe contrasts into the finest-scale high-pass band. The results in Section~\ref{sec:haar-transform} guarantee that the complete decomposition preserves total energy. What changes is how that energy is distributed across scales.

\section{Type b versus Type c Haar Inputs}
\label{app:type-bc}

\paragraph{Setup.}
We compare two PEGFAN input settings for DAHF's complete Haar transform~\citep{li2024permutation}. Type~b uses the normalized node features \(Z=X\), and Type~c uses the once-propagated features \(Z=\widehat A X\). Both settings retain the same four classifier channels \(X,\widehat A X,\widehat A^2X,\widehat A^3X\), with the same dataset-specific self-loop settings. We use the main experiment's \(R=7\) and dataset-specific \((h,\tau)\), the same direction dictionary and partition tree built from \(X\), and the same classifier and training settings. We train Type~b from scratch and reuse the main experiment's Type~c results. Both select checkpoints by validation loss and share ten fixed splits per dataset with seed \(42\), for 90 paired comparisons.

\begin{table}[H]
    \centering
    \caption{\textbf{Haar input ablation.} Test scores are means \(\pm\) sample standard deviations across ten splits (\%). For each split, \(\Delta\) is Type~b minus Type~c. The table reports the mean and sample standard deviation of these paired differences in percentage points. The final row gives equal weight to each dataset. The metrics are shown in separate panels. Teal marks average rows.}
    \label{tab:type-bc}

    \begin{pedafappendixtable}
\begin{tabular}{@{}l*{3}{>{\raggedleft\arraybackslash}m{95pt}}@{}}
\toprule
\rowcolor{white}
\multicolumn{4}{c}{\bfseries\itshape Accuracy} \\
\midrule
\rowcolor{white}
\textbf{Dataset} & \textbf{Type c} & \textbf{Type b} & \textbf{\(\Delta\)} \\
\midrule
\rowcolor{white}
Cora & 87.2233\,\ensuremath{\pm}\,1.2734 & 86.7002\,\ensuremath{\pm}\,1.5483 & \ensuremath{-}0.5231\,\ensuremath{\pm}\,0.9912 \\
\rowcolor{black!2}
CiteSeer & 77.5073\,\ensuremath{\pm}\,1.5369 & 76.8338\,\ensuremath{\pm}\,1.5031 & \ensuremath{-}0.6735\,\ensuremath{\pm}\,0.8463 \\
\rowcolor{white}
PubMed & 89.7541\,\ensuremath{\pm}\,0.3348 & 89.8276\,\ensuremath{\pm}\,0.3009 & +0.0735\,\ensuremath{\pm}\,0.1542 \\
\rowcolor{black!2}
Texas & 85.4054\,\ensuremath{\pm}\,4.9672 & 85.1351\,\ensuremath{\pm}\,5.2916 & \ensuremath{-}0.2703\,\ensuremath{\pm}\,2.6876 \\
\rowcolor{white}
Wisconsin & 87.6471\,\ensuremath{\pm}\,3.8167 & 86.0784\,\ensuremath{\pm}\,5.5034 & \ensuremath{-}1.5686\,\ensuremath{\pm}\,3.6741 \\
\rowcolor{black!2}
Cornell & 77.8378\,\ensuremath{\pm}\,3.9885 & 78.3784\,\ensuremath{\pm}\,2.8489 & +0.5405\,\ensuremath{\pm}\,2.1319 \\
\rowcolor{white}
Actor & 35.2303\,\ensuremath{\pm}\,1.3329 & 34.9342\,\ensuremath{\pm}\,1.0476 & \ensuremath{-}0.2961\,\ensuremath{\pm}\,1.0136 \\
\rowcolor{black!2}
Chameleon-filtered & 41.3061\,\ensuremath{\pm}\,3.0726 & 36.8758\,\ensuremath{\pm}\,2.7246 & \ensuremath{-}4.4303\,\ensuremath{\pm}\,2.6346 \\
\rowcolor{white}
Squirrel-filtered & 39.9151\,\ensuremath{\pm}\,1.7692 & 36.9494\,\ensuremath{\pm}\,2.4966 & \ensuremath{-}2.9656\,\ensuremath{\pm}\,1.7478 \\
\midrule
\rowcolor{PedafTableTeal}
\textbf{Avg.} & 69.0918 & 67.9681 & \ensuremath{-}1.1237 \\
\bottomrule
\end{tabular}
\par\medskip
\begin{tabular}{@{}l*{3}{>{\raggedleft\arraybackslash}m{95pt}}@{}}
\toprule
\rowcolor{white}
\multicolumn{4}{c}{\bfseries\itshape Macro-F1} \\
\midrule
\rowcolor{white}
\textbf{Dataset} & \textbf{Type c} & \textbf{Type b} & \textbf{\(\Delta\)} \\
\midrule
\rowcolor{white}
Cora & 85.9648\,\ensuremath{\pm}\,1.5814 & 85.4482\,\ensuremath{\pm}\,1.9957 & \ensuremath{-}0.5165\,\ensuremath{\pm}\,1.0365 \\
\rowcolor{black!2}
CiteSeer & 73.1023\,\ensuremath{\pm}\,2.2405 & 72.3641\,\ensuremath{\pm}\,1.8673 & \ensuremath{-}0.7382\,\ensuremath{\pm}\,1.8064 \\
\rowcolor{white}
PubMed & 89.3910\,\ensuremath{\pm}\,0.4493 & 89.4423\,\ensuremath{\pm}\,0.4175 & +0.0513\,\ensuremath{\pm}\,0.1770 \\
\rowcolor{black!2}
Texas & 60.8756\,\ensuremath{\pm}\,6.9947 & 61.2554\,\ensuremath{\pm}\,7.2708 & +0.3799\,\ensuremath{\pm}\,1.9051 \\
\rowcolor{white}
Wisconsin & 69.7074\,\ensuremath{\pm}\,10.4779 & 67.5847\,\ensuremath{\pm}\,10.2506 & \ensuremath{-}2.1228\,\ensuremath{\pm}\,6.1827 \\
\rowcolor{black!2}
Cornell & 60.3075\,\ensuremath{\pm}\,9.5616 & 62.4178\,\ensuremath{\pm}\,6.7369 & +2.1102\,\ensuremath{\pm}\,6.7179 \\
\rowcolor{white}
Actor & 27.0597\,\ensuremath{\pm}\,2.4644 & 27.6656\,\ensuremath{\pm}\,1.7647 & +0.6059\,\ensuremath{\pm}\,1.2392 \\
\rowcolor{black!2}
Chameleon-filtered & 34.3638\,\ensuremath{\pm}\,3.4473 & 27.5214\,\ensuremath{\pm}\,2.7393 & \ensuremath{-}6.8424\,\ensuremath{\pm}\,3.4365 \\
\rowcolor{white}
Squirrel-filtered & 27.4129\,\ensuremath{\pm}\,1.7805 & 24.0759\,\ensuremath{\pm}\,2.1649 & \ensuremath{-}3.3370\,\ensuremath{\pm}\,2.3771 \\
\midrule
\rowcolor{PedafTableTeal}
\textbf{Avg.} & 58.6872 & 57.5306 & \ensuremath{-}1.1566 \\
\bottomrule
\end{tabular}
\end{pedafappendixtable}
\end{table}

\paragraph{Results.}
Across the nine datasets, Type~c attains higher mean accuracy by \(1.1237\) percentage points and higher mean Macro-F1 by \(1.1566\) points. Type~c has higher observed accuracy on seven datasets and higher Macro-F1 on five. The largest gains are on Chameleon-filtered (\(4.43\) accuracy points and \(6.84\) Macro-F1 points) and Squirrel-filtered (\(2.97\) and \(3.34\) points). Across the other seven datasets, the corresponding mean gains are \(0.39\) and \(0.03\) points. Type~b scores higher on both metrics on Cornell, and the two settings score similarly on PubMed.

\section{Haar Support on Benchmark Graphs}
\label{app:support-audit}

We compare the support of \(W_\ell^\top W_\ell\delta_v\) at the same anchor \(v\) under the saved Paris-PEGFAN and direction-adaptive trees. Here \(\delta_v\) is the unit impulse at node \(v\). Both trees use complete Haar. This audit uses saved split-0 trees from the earlier \(R=7\) sweep. With seed \(42\), we sample up to 256 anchors per graph. For each anchor we match levels with 12--48 support nodes on each side and a larger-to-smaller support size ratio of at most \(1.3\). We retain 632 pairs, including 37 equal-size pairs across five datasets. Anchors can share support nodes, so we report descriptive paired statistics.

We measure density and the fraction of nodes of degree at least three on the subgraph induced by the support. We compute unweighted shortest-path distances when both induced subgraphs are connected. For directional concentration, we use the same frozen dictionary and row-normalized features on both sides. We compute the score once with equally weighted original nodes and once with child feature means weighted by their original node counts. We omit a concentration comparison when the corresponding feature scatter is at or below the threshold in the directional score.

\begin{table}[H]
\centering
\caption{Mean paired changes in Haar support from Paris-PEGFAN to direction-adaptive grouping. Density, branching, and the two concentration scores are in percentage points. Mean distance and diameter are in hops. \(n\) is the number of matched anchors. Light blue marks Texas, whose topology changes in the opposite direction.}
\label{tab:support-audit}

\begin{pedafappendixtable}
\begin{tabular}{@{}lrrrrrrr@{}}
\toprule
\rowcolor{white}
\textbf{Dataset} & \textbf{\(n\)} & \textbf{Density} & \textbf{Branching} & \textbf{Mean distance} & \textbf{Diameter} & \textbf{Node conc.} & \textbf{Child conc.} \\
\midrule
\rowcolor{black!6}
    \multicolumn{8}{@{}l}{\textbf{\textit{Comparable support sizes}}} \\
\rowcolor{white}
Cora & 110 & -0.888 & -6.431 & +0.637 & +1.836 & +0.836 & -0.626 \\
\rowcolor{black!2}
CiteSeer & 57 & -0.417 & -0.580 & +0.601 & +1.456 & +0.258 & +0.440 \\
\rowcolor{white}
PubMed & 83 & -0.274 & -2.171 & +0.070 & +0.145 & -0.007 & -0.654 \\
\rowcolor{PedafTableBlue!45}
Texas & 47 & +1.197 & +8.848 & -0.117 & -0.532 & +9.615 & +3.995 \\
\rowcolor{white}
Wisconsin & 137 & -0.628 & -3.831 & +0.598 & +1.861 & -0.957 & +4.552 \\
\rowcolor{black!2}
Cornell & 50 & -1.058 & -10.891 & +0.670 & +2.060 & +2.300 & +5.188 \\
\rowcolor{white}
Actor & 125 & -0.995 & -8.479 & +0.552 & +1.544 & +1.339 & +0.702 \\
\rowcolor{black!2}
Chameleon-filtered & 16 & -3.124 & -23.371 & +0.431 & +1.000 & -0.554 & -0.392 \\
\rowcolor{white}
Squirrel-filtered & 7 & -3.151 & -10.779 & +1.341 & +3.429 & -2.636 & -0.552 \\
    \midrule
\rowcolor{black!6}
    \multicolumn{8}{@{}l}{\textbf{\textit{Exactly equal support sizes}}} \\
\rowcolor{black!2}
Cora & 7 & -0.214 & -3.915 & +0.564 & +1.429 & -1.243 & -1.091 \\
\rowcolor{white}
CiteSeer & 7 & -0.470 & -2.821 & +0.424 & +1.000 & +0.323 & +0.011 \\
\rowcolor{black!2}
PubMed & 13 & -0.159 & -1.928 & +0.132 & +0.154 & +0.069 & -1.231 \\
\rowcolor{white}
Actor & 8 & -1.368 & -12.054 & +0.550 & +1.500 & +4.266 & +1.827 \\
\rowcolor{black!2}
Squirrel-filtered & 2 & -0.403 & -3.125 & +0.351 & +1.000 & -0.009 & -0.001 \\
\bottomrule
\end{tabular}
\end{pedafappendixtable}
\end{table}

In eight of the nine datasets, the size-matched means show lower density and branching and longer distances and diameters. Texas shows the reverse topology pattern. The five datasets with exact size matches show the same topology pattern as the eight, though the subset is small. Each concentration measure increases on five of the nine datasets, with both increasing on CiteSeer, Texas, Cornell, and Actor. Chameleon-filtered and Squirrel-filtered contribute only 16 and 7 matched anchors, respectively.

\begin{figure}[H]
    \centering
    \includegraphics[width=.49\linewidth]{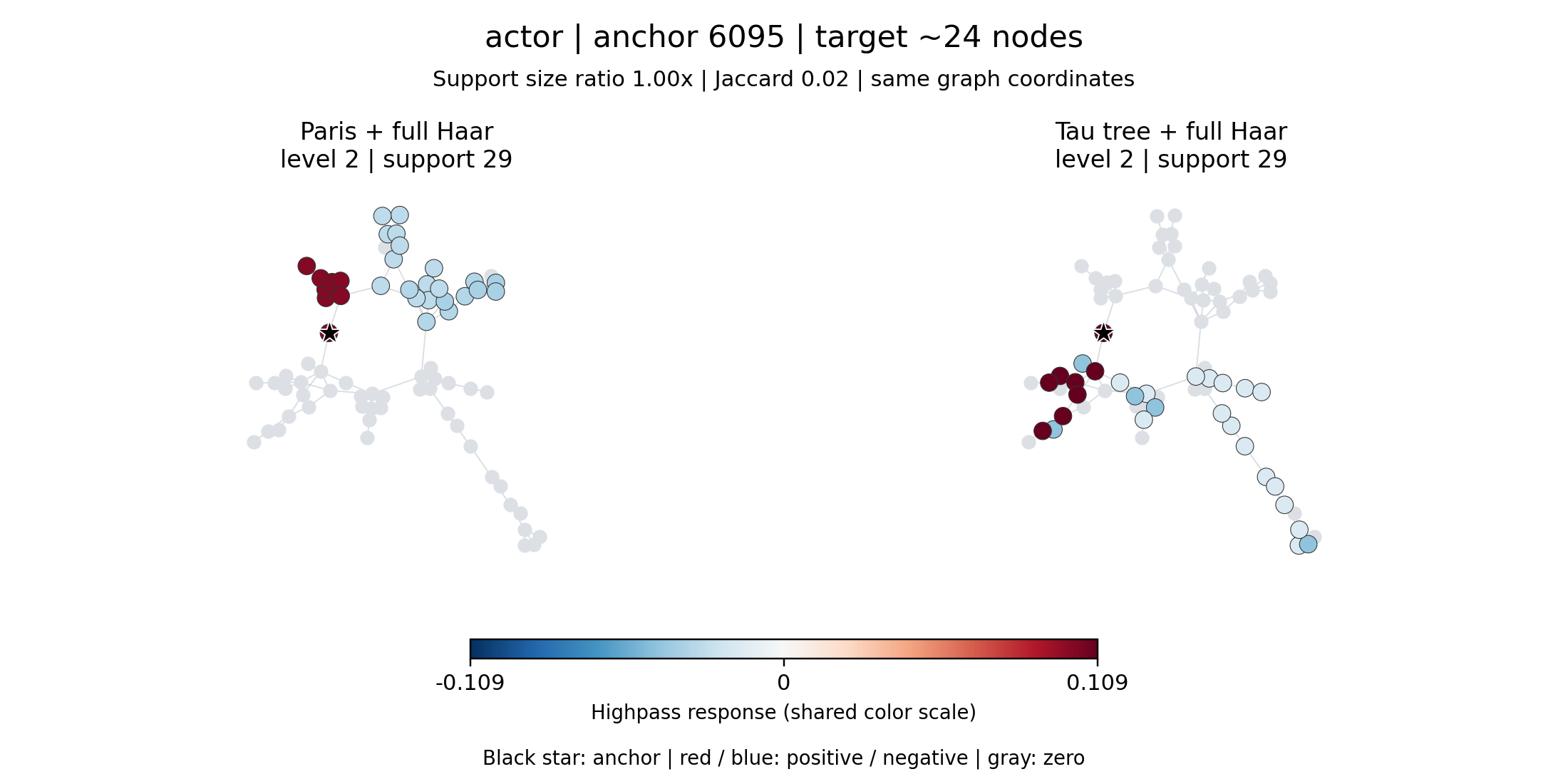}\hfill
    \includegraphics[width=.49\linewidth]{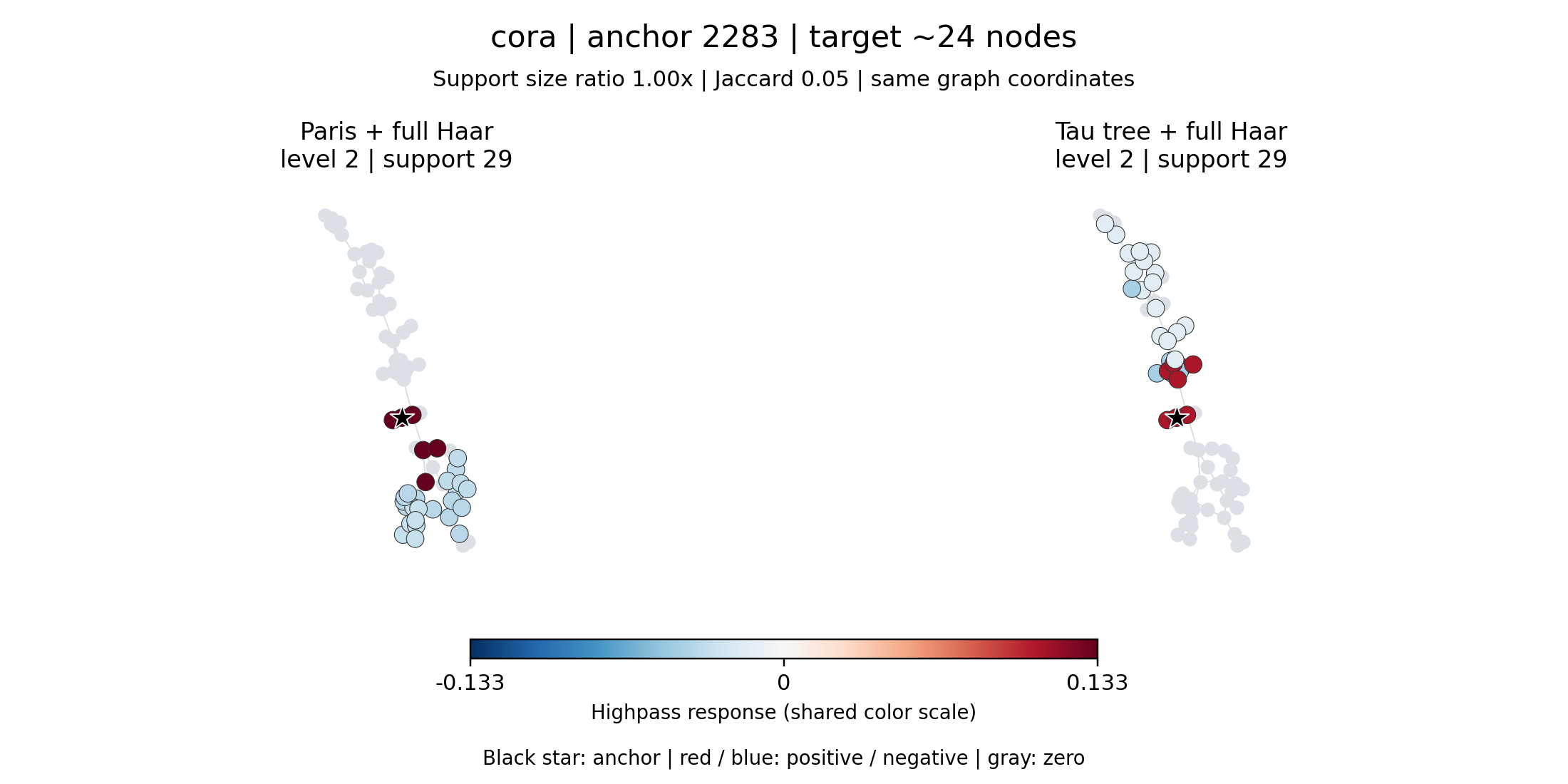}
    \caption{Two illustrative impulse-response supports, Actor node 6095 (left) and Cora node 2283 (right). Each case has 29 support nodes at level 2 in both trees. On Actor, the induced edge count changes from 33 to 28 and diameter from 8 to 12 hops. On Cora, the edge count changes from 36 to 42 and diameter from 8 to 11 hops. The two trees within each case use the same node layout and response color scale.}
    \label{fig:support-audit}
\end{figure}

\section{Greedy Grouping versus Local Exhaustive Search}
\label{app:greedy-audit}

We sample 120 distinct connected 11-node neighborhoods from each benchmark graph using seed \(42\), fit the \(R=7\) dictionary on the full graph, and compare the first greedy parent on each induced neighborhood with the highest-scoring eight-node parent reachable by any seed and growth sequence that passes the same structural screen. We use \(h=8\) and \(\tau=0.5\) throughout. This search optimizes the score \(C\) of one parent on a small subgraph. No classifiers are trained for this check.

\begin{table}[H]
\centering
\caption{Directional score \(C\) for greedy grouping and exhaustive search. Counts include all 120 neighborhoods per dataset. The gap is the exhaustive score minus the greedy score, in units of \(C\).}
\label{tab:greedy-audit}

\begin{pedafappendixtable}
\begin{tabular}{@{}lrrrrrrr@{}}
\toprule
\rowcolor{white}
\textbf{Dataset} & \textbf{Greedy} & \textbf{Exhaustive} & \textbf{Mean gap} & \textbf{Max.\ gap} & \textbf{Improved} & \textbf{Gap \(\ge 0.02\)} & \textbf{Gap \(\ge 0.05\)} \\
\midrule
\rowcolor{white}
Cora & 0.0601 & 0.0634 & 0.0034 & 0.0736 & 67 & 4 & 1 \\
\rowcolor{black!2}
CiteSeer & 0.0244 & 0.0259 & 0.0015 & 0.0112 & 69 & 0 & 0 \\
\rowcolor{white}
PubMed & 0.0440 & 0.0475 & 0.0035 & 0.0462 & 55 & 5 & 0 \\
\rowcolor{black!2}
Texas & 0.1834 & 0.1913 & 0.0079 & 0.1928 & 44 & 14 & 4 \\
\rowcolor{white}
Wisconsin & 0.1590 & 0.1716 & 0.0125 & 0.1243 & 58 & 26 & 8 \\
\rowcolor{black!2}
Cornell & 0.1547 & 0.1663 & 0.0115 & 0.1250 & 59 & 24 & 9 \\
\rowcolor{white}
Actor & 0.2911 & 0.3126 & 0.0215 & 0.3924 & 64 & 33 & 15 \\
\rowcolor{black!2}
Chameleon-filtered & 0.0372 & 0.0411 & 0.0039 & 0.0903 & 56 & 8 & 2 \\
\rowcolor{white}
Squirrel-filtered & 0.0578 & 0.0674 & 0.0096 & 0.3831 & 62 & 13 & 4 \\
\bottomrule
\end{tabular}
\end{pedafappendixtable}
\end{table}

The first greedy parent reaches eight nodes in all 1,080 neighborhoods. Exhaustive search improves the score in 534 cases, with a mean gap of \(0.0084\). There are 127 gaps of at least \(0.02\) and 43 of at least \(0.05\). Actor has the largest mean gap (\(0.0215\)). The largest individual gaps on Actor and Squirrel-filtered reach \(0.3924\) and \(0.3831\), respectively.

\end{document}